\documentclass[journal]{IEEEtran}
\usepackage{amsmath,amssymb}
\usepackage[scr=stixfancy]{mathalpha}
\usepackage{algorithmic}
\usepackage{algorithm}
\usepackage{booktabs}
\usepackage{multirow}
\usepackage{graphicx}
\usepackage[noadjust]{cite}
\usepackage{stfloats}
\usepackage{needspace}
\usepackage{xcolor}
\usepackage[hidelinks]{hyperref}
\hypersetup{pdfauthor={Haitong Wang, Souren Pashangpour, Matthew Lisondra, Yaseen Rehman, Goldie Nejat}, pdftitle={ViSiR-Parkour: Visual Sim-to-Real Locomotion for Scalable and Generalizable Legged Robot In-the-Wild Parkour}}

\makeatletter
\def\@IEEEtablecaptionsepspace{\vskip0pt\relax}
\makeatother

\title{ViSiR-Parkour: Visual Sim-to-Real Locomotion for Scalable and Generalizable Legged Robot In-the-Wild Parkour}

\author{Haitong~Wang, Souren~Pashangpour\textsuperscript{*}, Matthew~Lisondra\textsuperscript{*}, Yaseen~Rehman, and~Goldie~Nejat, \textit{IEEE Member}%
\thanks{This work has been submitted to the IEEE for possible publication. Copyright may be transferred without notice, after which this version may no longer be accessible.}%
\thanks{This work was supported by the Natural Sciences and Engineering Research Council of Canada (NSERC), the Innovation, Equity \& the Future of Prosperity (IEP) program of the Canadian Institute for Advanced Research (CIFAR), and the Canada Research Chairs Program. *These authors contributed equally to this work. \textit{Corresponding author: Haitong Wang.}}%
\thanks{The authors are with the Autonomous Systems and Biomechatronics Laboratory (ASBLab), Department of Mechanical and Industrial Engineering, University of Toronto, Toronto, ON M5S 3G8, Canada (e-mail: \href{mailto:haitong.wang@mail.utoronto.ca}{haitong.wang@mail.utoronto.ca}; \href{mailto:souren.pashangpour@mail.utoronto.ca}{souren.pashangpour@mail.utoronto.ca}; \href{mailto:matthew.lisondra@mail.utoronto.ca}{matthew.lisondra@mail.utoronto.ca}; \href{mailto:yaseen.rehman@mail.utoronto.ca}{yaseen.rehman@mail.utoronto.ca}; \href{mailto:goldie.nejat@utoronto.ca}{goldie.nejat@utoronto.ca}).}}

\begin{document}
\bstctlcite{IEEEReferenceAuthorControl}
\maketitle
\raggedbottom

\begin{abstract}
\noindent Visual locomotion methods for legged robot parkour face challenges in transferring policies from simulations to real-world environments as well as generalizing to unseen terrain geometries and layouts. In this paper, we introduce Visual Sim-to-Real Parkour (ViSiR-Parkour), a novel architecture for generalizable legged robot parkour in the wild. ViSiR-Parkour trains a single teacher policy with privileged observations on a scaled-up distribution of diverse primitive and combination terrains. A unique progress-aware terrain reallocation (PATR) curriculum learning approach uses traversal success rates to reallocate training robots to terrain types requiring further learning. A visual student policy is then trained using hybrid policy learning with teacher supervision to acquire locomotion skills and task reward optimization to improve terrain traversal beyond imitating teacher actions. Our visual sim-to-real method is the first to apply learning-based stereo matching to legged robot parkour, transforming onboard stereo observations into clean metric depth images consistent with the simulated visual inputs obtained during training. ViSiR-Parkour enables direct policy transfer without depth-noise augmentation or real-world policy fine-tuning. Simulated and real-world experiments demonstrate the performance and generalizability of ViSiR-Parkour across diverse challenging indoor and outdoor terrains. A scalability study shows generalizability to unseen out-of-distribution terrains improves as the number of training terrain types increases. An ablation study validates the main design choices of the architecture. Project page: \url{https://visir-parkour.github.io}.
\end{abstract}


\section{Introduction}
\noindent \IEEEPARstart{L}{EGGED} robot locomotion enables robots to traverse uneven terrains including stairs, slopes, and rubble \cite{learningAgileDynamicMotorSkills}. This capability is crucial for real-world tasks such as navigation \cite{handDrawnMapNavigation,xNav,egoNav}, exploration of unknown environments \cite{mineTunnelExploration,traversabilityAwareExploration}, and mobile manipulation \cite{umiOnLegs,humanPlus,embodiedAISurvey}. In particular, robot parkour requires legged robots to engage in dynamic, agile maneuvers over challenging, irregular, and discrete terrains such as jumping across terrain gaps, crossing narrow beams, and climbing rock piles \cite{anymalParkour,robotParkourLearning,parkourInTheWild,agilePerceptiveMultiskillLocomotion}. In order to achieve this, locomotion policies are often trained in simulation using deep reinforcement learning (DRL) \cite{PPO} and then deployed on physical robots in real environments \cite{learningAgileDynamicMotorSkills}.
\nocite{extremeParkour} %

Sim-to-real transfer is still challenging as simulation cannot accurately reproduce how a physical robot directly interacts with its environment \cite{dynamicsRandomization}. Existing sim-to-real methods for legged robots have primarily focused on gaps in robot dynamics which include mismatches between simulation and reality in the robot's physical properties, actuator behaviors, and contact with the environment \cite{PACE}. These methods reduce the dynamics gap by accurately modeling or estimating robot mass, actuator response, joint friction, and/or foot--ground contact \cite{learningAgileDynamicMotorSkills,challengingTerrainLocomotion,dynamicsRandomizationRevisited,locoFormer,simToRealQuadrupedLocomotion,beyondMimic,ASAP,athleticLocoManipulationSimToReal,RMA,RL2AC}.

Another sim-to-real challenge for legged robots is the visual sim-to-real gap where discrepancies between simulation-rendered images and real-world robot camera images exist \cite{nowYouSeeThat,hikingInTheWild}. Depth-based visual sim-to-real methods primarily use domain randomization (DR) \cite{domainRandomization} to apply data augmentation, such as adding edge noise and Gaussian blur \cite{parkourInTheWild} to simulated depth images, so that trained policies can be deployed using raw noisy depth images from onboard real-world cameras \cite{robotParkourLearning,parkourInTheWild,extremeParkour,nowYouSeeThat,hikingInTheWild,visualLocomotion,ladderMan}. RGB-based methods, on the other hand, apply DR to increase appearance diversity through image generation \cite{visualParkourGeneratedImages} or large-scale randomization over lighting, textures, and camera parameters \cite{VIRAL,doorMan}. However, for depth-based methods, applying heavy domain randomization can cause the locomotion policy to be over-conservative, as the policy receives degraded visual observations and therefore learns to take restrained actions to avoid failures caused by perceptual uncertainty \cite{dynamicsRandomizationRevisited}. RGB-based methods require computationally expensive rendering during training \cite{visualParkourGeneratedImages} and large-scale randomization \cite{VIRAL,doorMan} for the policy to transfer to the real world. Furthermore, RGB images do not directly provide metric terrain geometry (e.g., obstacle height and gap width), which is important for legged robots to select safe footholds during parkour.

Deploying visual locomotion policies in the wild also requires terrain generalization in order to handle terrains with geometries and layouts not encountered during training. Prior locomotion policies are often trained on small, fixed terrain sets (e.g., stairs and slopes) \cite{agilePerceptiveMultiskillLocomotion,hikingInTheWild,PIE,START}, or a single terrain type (e.g., ladders and stepping beams) \cite{robustLadderClimbing,stairMaster,walkThePLANC}. Such limited terrain coverage in training may not be able to extend the locomotion policy to safely navigate terrains with unseen geometries or layouts.

Several recent works have considered terrain generalization to: 1) learn terrain-map representations using attention-based map encoding (AME) \cite{AME,AME2}; 2) expand terrain coverage during policy training or adaptation \cite{parkourInTheWild,TTTParkour}; or 3) automatically generate diverse terrains in simulation \cite{terrainRobustnessBenchmark,diverseChallengingTerrains}. As AME-based methods require elevation mapping, they cannot directly apply to visual parkour as they condition the locomotion policy on a constructed robot-centric terrain height map rather than directly on onboard visual observations. The coverage-expansion methods require recursive finetuning on new terrains \cite{parkourInTheWild}, or reconstructing the target terrain in simulation to finetune the policy at test time \cite{TTTParkour}. As a result, these methods cannot be directly deployed on unseen terrains without additional fine-tuning. Lastly, terrain generation methods primarily produce continuous height-field surfaces, limiting training exposure to discrete terrains such as stepping stones and rock piles \cite{terrainRobustnessBenchmark,diverseChallengingTerrains}.

In this paper, to address the aforementioned challenges of both the visual sim-to-real gap and in-the-wild terrain generalization, we propose Visual Sim-to-Real Parkour (ViSiR-Parkour), a novel legged robot architecture that combines learning-based stereo matching \cite{LAS2}, scaled-up terrain diversity, and a progress-aware terrain reallocation (PATR) curriculum for in the wild parkour. The main contributions of this paper are:

1) The introduction of a visual sim-to-real method for legged robot parkour that is the first to use a learning-based stereo matching model to generate accurate clean depth images from real-world stereo observations. Our sim-to-real method does not require depth-noise augmentation, and uniquely reconstructs metric depth from real-world stereo image pairs. Real-world observations are aligned with the clean depth representation used during simulation training, enabling direct sim-to-real transfer of the parkour policy.

2) The development of a novel PATR curriculum learning strategy that computes a training-priority score for each terrain type from its terrain traversal success rates, then reallocates parallel training robots across terrain types according to the resulting scores. This curriculum learning strategy allows for training a single policy across diverse terrain types. We therefore scale up terrain diversity for legged robot parkour training by expanding both the range of terrain types and the geometric and spatial variations within each type, enabling zero-shot generalization to unseen in-the-wild terrains.

\section{Related Works}
We discuss the following pertinent literature on legged robots with respect to: 1) sim-to-real strategies \cite{learningAgileDynamicMotorSkills,robotParkourLearning,parkourInTheWild,extremeParkour,PACE,challengingTerrainLocomotion,dynamicsRandomizationRevisited,locoFormer,simToRealQuadrupedLocomotion,beyondMimic,ASAP,athleticLocoManipulationSimToReal,RMA,RL2AC,nowYouSeeThat,hikingInTheWild,visualLocomotion,ladderMan,visualParkourGeneratedImages,VIRAL,doorMan}; and 2) in-the-wild terrain generalization \cite{parkourInTheWild,AME,AME2,TTTParkour,terrainRobustnessBenchmark,diverseChallengingTerrains}.

\subsection{Sim-to-Real Strategies}
Two types of sim-to-real strategies for legged robots are discussed here: 1) dynamics, and 2) visual.

\subsubsection{Dynamics Sim-to-Real}
Dynamics sim-to-real methods reduce the gap between simulation and real-world deployment caused by inaccurate modeling of the robot and its interaction with the environment. This gap can come from rigid body parameters, actuator response, contact dynamics and payload changes. Existing methods have mainly used DR \cite{challengingTerrainLocomotion,dynamicsRandomizationRevisited,locoFormer} and/or system identification (SysID) \cite{learningAgileDynamicMotorSkills,PACE,simToRealQuadrupedLocomotion,beyondMimic,ASAP,athleticLocoManipulationSimToReal,RMA,RL2AC}.

DR methods randomize simulator parameters during training in order for the physical robot dynamics to be within the policy's training distribution \cite{domainRandomization}. These randomized parameters include robot physical properties such as mass, inertia, morphology, and payload \cite{dynamicsRandomizationRevisited,locoFormer}; actuator and controller properties such as PD gains and motor delay \cite{dynamicsRandomizationRevisited,locoFormer}; and robot-environment interaction properties, such as foot-terrain friction, external disturbances, proprioception noise \cite{challengingTerrainLocomotion,locoFormer}. These DR methods have been used for quadruped locomotion over challenging terrains \cite{challengingTerrainLocomotion,dynamicsRandomizationRevisited}, and cross-embodiment locomotion with various quadrupeds, humanoids and wheeled-legged quadrupeds \cite{locoFormer}.

SysID methods can be further categorized into accurate robot modeling \cite{simToRealQuadrupedLocomotion,beyondMimic}, offline SysID \cite{learningAgileDynamicMotorSkills,PACE,ASAP,athleticLocoManipulationSimToReal}, and online SysID methods \cite{RMA,RL2AC}. SysID methods with accurate modeling use measured robot descriptions, actuator models, latency models, and inertial properties such as joint armature to design the simulated robot dynamics close to the physical robot \cite{simToRealQuadrupedLocomotion,beyondMimic}. Offline SysID methods collect real-world robot data to either: 1) train an actuator network to predict the joint output torque in simulation \cite{learningAgileDynamicMotorSkills}, 2) fit actuator dynamics parameters using evolutionary parameter search optimization \cite{PACE}, or 3) train residual dynamics models using DRL to predict corrective robot actions \cite{ASAP} or actuator torques \cite{athleticLocoManipulationSimToReal} in simulation. Online SysID methods estimate dynamics information during deployment from recent robot states, either as a latent environment representation \cite{RMA} or as adaptive torque compensation \cite{RL2AC} to address the sim-to-real mismatch.
\subsubsection{Visual Sim-to-Real}
Visual sim-to-real methods reduce the visual perception gap between simulation and the real world by aligning simulated camera observations with real-world onboard camera observations \cite{robotParkourLearning,parkourInTheWild,extremeParkour,nowYouSeeThat,hikingInTheWild,visualLocomotion,visualParkourGeneratedImages,VIRAL,doorMan}. These methods can be categorized based on the visual input modality used into depth-based methods \cite{robotParkourLearning,parkourInTheWild,extremeParkour,nowYouSeeThat,hikingInTheWild,visualLocomotion,ladderMan}, and RGB-based methods \cite{visualParkourGeneratedImages,VIRAL,doorMan}. Depth-based methods mainly apply DR to model the real-world depth-camera artifacts. For example, some methods model generic depth artifacts by injecting depth noise, missing pixels, holes, and edge artifacts into the simulated depth images, \cite{robotParkourLearning,parkourInTheWild,visualLocomotion}. Other methods further model stereo camera depth artifacts, including stereo matching errors, and calibration errors \cite{nowYouSeeThat,hikingInTheWild}. A handful of methods apply depth clipping, filtering, and camera latency during training \cite{robotParkourLearning,parkourInTheWild,extremeParkour,hikingInTheWild}. In \cite{ladderMan}, LadderMan reduces the visual sim-to-real gap by using a stereo matching model to generate clean depth images for perceptive humanoid ladder climbing. Multiple expert policies are trained through DRL for different ladder configurations using human ladder-climbing motion data and target contact positions generated by geometry-based heuristics that assign predefined hand and foot locations on ladder rungs. These expert policies are distilled into a visual ladder climbing policy. In general, these depth-based visual sim-to-real methods have been used for quadruped locomotion \cite{robotParkourLearning,parkourInTheWild,extremeParkour,visualLocomotion}, and humanoid locomotion \cite{nowYouSeeThat,hikingInTheWild,ladderMan}.

RGB-based methods also apply DR by increasing the appearance diversity and image fidelity instead of modeling artifacts like depth-based methods \cite{visualParkourGeneratedImages,VIRAL,doorMan}. For example, in \cite{visualParkourGeneratedImages}, simulated depth images and semantic masks are used to generate egocentric RGB observations, and optical-flow warping is used to produce temporally consistent RGB sequences for training. VIRAL (Visual Sim-to-Real for Humanoid Loco-Manipulation) \cite{VIRAL} and DoorMan \cite{doorMan} randomize scene illumination and surface textures, as well as camera pose, field of view, intrinsics, and sensor delay. These RGB-based visual sim-to-real methods have been used for quadruped parkour \cite{visualParkourGeneratedImages}, and humanoid loco-manipulation \cite{VIRAL,doorMan}.

\subsection{In-the-Wild Terrain Generalization}
Prior terrain generalization methods for legged robot locomotion use: 1) terrain-map representations \cite{AME,AME2}, 2) terrain-coverage expansion \cite{parkourInTheWild,TTTParkour}, or 3) automatic terrain generation \cite{terrainRobustnessBenchmark,diverseChallengingTerrains}. In \cite{AME}, AME encodes a robot-centric terrain height map for policy learning by extracting point-wise map features using a convolutional neural network (CNN), and applying attention conditioned on proprioception to weigh terrain points for locomotion control. In \cite{AME2}, AME-2 extends the map encoding method with local and global elevation map features and a learning-based elevation mapping pipeline. AME-based methods have been applied to legged robot parkour on challenging terrains in both indoor and outdoor environments \cite{AME,AME2}.

In \cite{parkourInTheWild}, the Parkour-in-the-Wild framework expands terrain coverage by training terrain-specific experts, distilling them into a single visual policy, and recursively finetuning the distilled policy on new terrains using DRL \cite{PPO}. The finetuned policy has been deployed on a quadruped for parkour over challenging indoor and outdoor terrains. In \cite{TTTParkour}, test-time training parkour (TTT-Parkour) was introduced to reconstruct real-world terrains into a simulation mesh and finetune a pretrained parkour policy on the reconstructed terrain mesh at test time. TTT-Parkour has been tested on a humanoid for indoor parkour.

Terrain generation methods expand simulated terrain coverage by using a conditional generative adversarial network (GAN) to generate realistic unstructured height maps \cite{terrainRobustnessBenchmark}, or by using quality-diversity search to generate height fields that expose a policy to locomotion failure modes \cite{diverseChallengingTerrains}. They have been used to fine-tune DRL locomotion policies and achieved robust terrain traversal on a biped and a quadruped in simulation \cite{diverseChallengingTerrains}.

\subsection{Summary of Limitations}
The majority of legged robot sim-to-real methods have focused on reducing dynamics mismatches through DR and SysID \cite{learningAgileDynamicMotorSkills,PACE,challengingTerrainLocomotion,dynamicsRandomizationRevisited,locoFormer,simToRealQuadrupedLocomotion,beyondMimic,ASAP,athleticLocoManipulationSimToReal,RMA,RL2AC}. However, sim-to-real transfer for perceptive locomotion also requires addressing visual discrepancies as locomotion policies rely on visual observations to adapt movements to challenging terrains \cite{parkourInTheWild,nowYouSeeThat}. For visual sim-to-real, depth-based methods can contain depth corruption which can obscure terrain geometry, causing the robot to misjudge where to step and fail to traverse parkour terrains \cite{robotParkourLearning,parkourInTheWild,extremeParkour,nowYouSeeThat,hikingInTheWild,visualLocomotion}. An exception is LadderMan \cite{ladderMan}, which uses a learning-based stereo matching model to obtain clean depth observations for humanoid ladder climbing. However, it relies on human ladder-climbing motion data and predefined target contact positions on ladder rungs. For irregular and discrete terrains, determining dynamically feasible contact position sequences using LadderMan's geometry-based heuristics is not feasible as contact targets and specific terrain geometry are not a priori known \cite{highSpeedControlNavigation,deepTrackingControl}.

RGB-based visual sim-to-real methods generate visual diversity which places heavy computational demands on policy training \cite{visualParkourGeneratedImages,VIRAL,doorMan}. Moreover, RGB images lack explicit metric terrain geometry (e.g., obstacle height), which is important for precise foot placement.

For terrain generalization, AME-based methods rely on elevation mapping, making locomotion performance dependent on the accuracy of the reconstructed terrain maps \cite{AME,AME2}. Coverage-expansion methods require repeated policy training as new or reconstructed terrains are added \cite{parkourInTheWild,TTTParkour}. Terrain generation methods focus on height variations in continuous-support terrains, without considering coverage of sparse-support terrains such as stepping stones and balance beams \cite{terrainRobustnessBenchmark,diverseChallengingTerrains}. This limits policy training for foothold selection for such terrains as well as traversing these separated support structures.

To address the aforementioned limitations of existing visual sim-to-real methods, our architecture uses a learning-based stereo matching model to obtain clean depth observations, preserving metric terrain geometry for precise foot placement without the need for depth-noise augmentation used in depth-based methods or the computational cost of appearance randomization in RGB-based methods. To enable in-the-wild terrain generalization, we develop the PATR curriculum learning method for training a single parkour policy across diverse terrain types, without relying on reference motion data or predefined target contact positions. We scale training terrain diversity to include a wider range of terrain types and geometric variations than any prior parkour method. Training on this scaled terrain distribution enables the policy to generalize to unseen terrains without additional fine-tuning.

\begin{figure}[!b]
\centering

\vspace{-7.5pt}
\includegraphics[width=0.95\columnwidth]{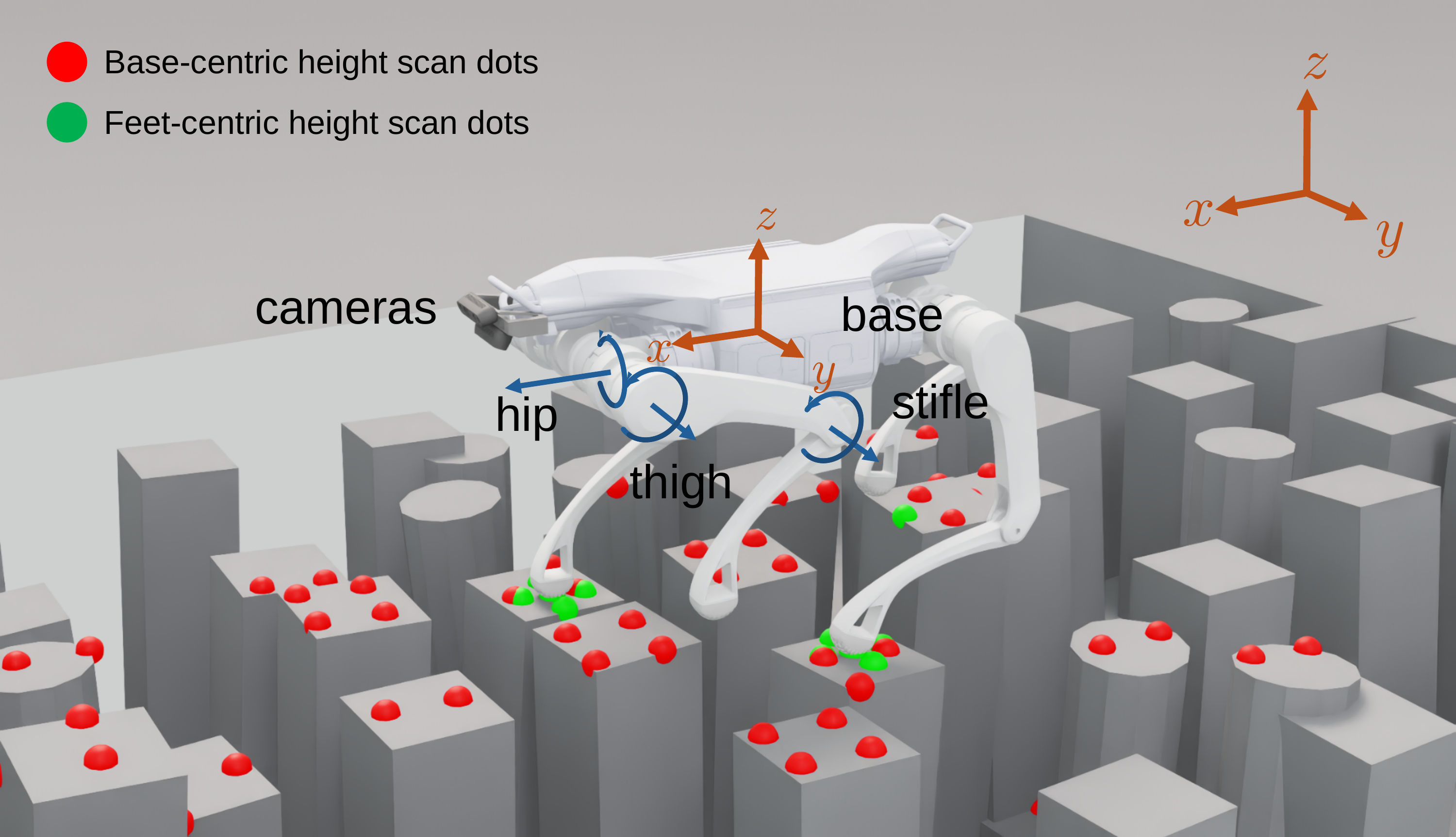}
\caption{Robot base frame, joints, and base-centric (red) and feet-centric (green) terrain height-scan dots, shown on an example stepping stones terrain.}
\label{fig:base_frame}
\end{figure}

\section{Problem Formulation}

The visual parkour problem requires learning a single policy $\pi_\theta$ that enables a legged robot $\mathscr{r}$ to follow a velocity command $\boldsymbol{v}^{\mathrm{cmd}}$ to traverse challenging terrains using onboard sensors, where $\boldsymbol{v}^{\mathrm{cmd}}=(v_x^{\mathrm{cmd}},v_y^{\mathrm{cmd}},\omega_z^{\mathrm{cmd}})\in\mathbb{R}^3$ represents the linear forward, lateral and angular command velocities. In this paper, we apply this problem to a quadruped with 12 degrees of freedom (DOFs), a widely used configuration for quadrupedal robots \cite{PACE}. The robot has both a forward-facing and a downward-facing stereo camera, a body-mounted IMU, and joint encoders. The robot base frame and positive leg-joint rotation directions are shown in Fig. \ref{fig:base_frame}. At each timestep $t$, the parkour policy $\pi_\theta$ generates an action $\boldsymbol{a}_t$ using the observation $\boldsymbol{o}_t$ and receives a reward $r_t$:

\begin{equation}
\boldsymbol{a}_t=\pi_\theta(\boldsymbol{o}_t),
\end{equation}
\noindent where $\boldsymbol{o}_t=(\boldsymbol{o}_{\mathrm{depth},t},\boldsymbol{v}^{\mathrm{cmd}}_t,\boldsymbol{o}_{\mathrm{prop},t})$ includes: 1) the visual observations represented by depth images $\boldsymbol{o}_{\mathrm{depth}}\in\mathbb{R}^{2\times H\times W}$ obtained from the two onboard stereo cameras, where $H$ and $W$ are the depth image height and width in pixels, respectively; 2) the externally specified velocity command $\boldsymbol{v}^{\mathrm{cmd}}$, and 3) the proprioceptive observations, $\boldsymbol{o}_{\mathrm{prop}}\in\mathbb{R}^{42}$. The proprioceptive observations include the robot base angular velocity $\boldsymbol{\omega}\in\mathbb{R}^3$ and the projected gravity $\boldsymbol{g}_t\in\mathbb{R}^3$, both obtained from the IMU; the joint positions $\boldsymbol{q}\in\mathbb{R}^{12}$ and joint velocities $\dot{\boldsymbol{q}}\in\mathbb{R}^{12}$ obtained from the joint encoders; and the robot's previous policy action $\boldsymbol{a}_{t-1}\in\mathbb{R}^{12}$.

The robot action $\boldsymbol{a}_t\in\mathbb{R}^{12}$ represents the normalized joint position offsets for the 12 actuated joints. The action is mapped to the desired joint positions $\boldsymbol{q}_t^{\mathrm{d}}$:
\begin{equation}
\boldsymbol{q}_t^{\mathrm{d}}=\boldsymbol{q}^{\mathrm{nom}}+\boldsymbol{s}_a\odot\boldsymbol{a}_t,
\end{equation}
\noindent where $\boldsymbol{q}^{\mathrm{nom}}\in\mathbb{R}^{12}$ represents the nominal joint positions, $\boldsymbol{s}_a\in\mathbb{R}^{12}$ is the action scale, and $\odot$ denotes element-wise multiplication. The desired joint positions are sent to a low-level robot Proportional-Derivative (PD) controller to execute the policy action by generating the joint motor torques required to track the desired joint positions:
\begin{equation}
\boldsymbol{\tau}_t=\boldsymbol{K}_p(\boldsymbol{q}_t^{\mathrm{d}}-\boldsymbol{q}_t)+\boldsymbol{K}_d\dot{\boldsymbol{q}}_t,
\end{equation}
\noindent where $\boldsymbol{\tau}_t$ denotes the joint motor torque, and $\boldsymbol{K}_p$, $\boldsymbol{K}_d$ denote the PD gains.

The objective is to learn policy parameters $\theta$ that maximize the expected discounted return:
\begin{equation}
\theta^*=\arg\max_\theta\mathbb{E}_{\pi_\theta}\left[\sum_{t=0}^{T-1}\gamma^t r_t\right],
\end{equation}
\noindent where $\gamma$ denotes the discount factor, and $T$ is the total number of timesteps in an episode. Table \ref{tab:nomenclature} summarizes the notation used throughout the paper.

\suppressfloats[t]
\begin{table}[!t]
\caption{Nomenclature}
\label{tab:nomenclature}
\centering
\footnotesize
\setlength{\tabcolsep}{3pt}
\begin{tabular}{@{}p{0.39\columnwidth}p{0.58\columnwidth}@{}}
\toprule
Notation & Description \\
\midrule
$\pi_\phi,\pi_\theta$ & Teacher/student policies \\
$\boldsymbol{o}_{\phi,t},\boldsymbol{o}_t$ & Teacher/student observations \\
$\boldsymbol{o}_{\mathrm{depth},t},\boldsymbol{o}_{\mathrm{prop},t}$ & Depth/proprioceptive observations \\
$\boldsymbol{o}_{\mathrm{scan},t}^{b},\boldsymbol{o}_{\mathrm{scan},t}^{f}$ & Base-/feet-centric terrain scans \\
$\boldsymbol{a}_t,\boldsymbol{a}^{\phi}_t$ & Student/teacher actions \\
$\boldsymbol{h}_t$ & GRU hidden state \\
$\boldsymbol{z}^{\mathrm{vis}}_t,\boldsymbol{z}^{\mathrm{fuse}}_t$ & Visual/fused features \\
$\boldsymbol{I}_t^c$ & Stereo image pair for camera $c$ \\
$g_k$ & Terrain training-priority score \\
$\ell_i,s_k$ & Stored home level; terrain-type success score \\
$\boldsymbol{v}_{xy}^{\mathrm{cmd}},\omega_z^{\mathrm{cmd}}$ & Commanded linear/yaw velocities \\
$v_{\mathrm{th}},\omega_{\mathrm{th}}$ & Forward / in-place rotation thresholds \\
$\boldsymbol{v}_{xy},v_z$ & Horizontal/vertical base linear velocities \\
$\boldsymbol{\omega}_{xy},\omega_z$ & Base roll/pitch and yaw angular velocities \\
$d_t$ & Base displacement along command \\
$\boldsymbol{q},\dot{\boldsymbol{q}},\ddot{\boldsymbol{q}}$ & Actual joint position/velocity/acceleration \\
$\boldsymbol{\tau}$ & Joint torques \\
$\boldsymbol{q}^{\mathrm{nom}},\boldsymbol{q}^{\mathrm{ref}},\boldsymbol{q}^{\mathrm{d}}_t$ & Nominal/reference/desired joint positions \\
$\underline{\boldsymbol{q}}^{\mathrm{lim}},\overline{\boldsymbol{q}}^{\mathrm{lim}},\dot{\boldsymbol{q}}^{\mathrm{lim}}$ & Soft joint position/velocity limits \\
$\mathcal{J}_{\mathrm{hip}},\mathcal{J}_{\mathrm{thigh,calf}}$ & Sets of hip joints and thigh/calf joints \\
$\boldsymbol{F}_i$ & Net contact force on robot link $i$ \\
$\mathcal{B}$ & Set of base, hip, thigh, and calf links \\
$\boldsymbol{v}_{f,xy}$ & Horizontal linear velocity of foot $f$ \\
$t_f^{\mathrm{air}}$ & Air time of foot $f$ before touchdown \\
$h_f,h_{f,k}$ & Foot clearance; scan height ($k{=}0$: center) \\
$\mathbf{1}_f^{\mathrm{land}},\mathbf{1}_f^{\mathrm{contact}}$ & Foot landing/contact indicators \\
$\mathbf{1}^{\mathrm{move}},\mathbf{1}^{\mathrm{stand}},\mathbf{1}^{\mathrm{fall}}$ & Moving/standing/base-contact indicators \\
$\mathbf{1}[\cdot]$ & Indicator function \\
$[x]_+$ & $\max(x,0)$ \\
$r_t$ & Reward at timestep $t$ \\
\bottomrule
\end{tabular}
\end{table}

\section{Visual Sim-to-Real Parkour Architecture}
\begin{figure*}[!t]
\centering
\includegraphics[width=\textwidth]{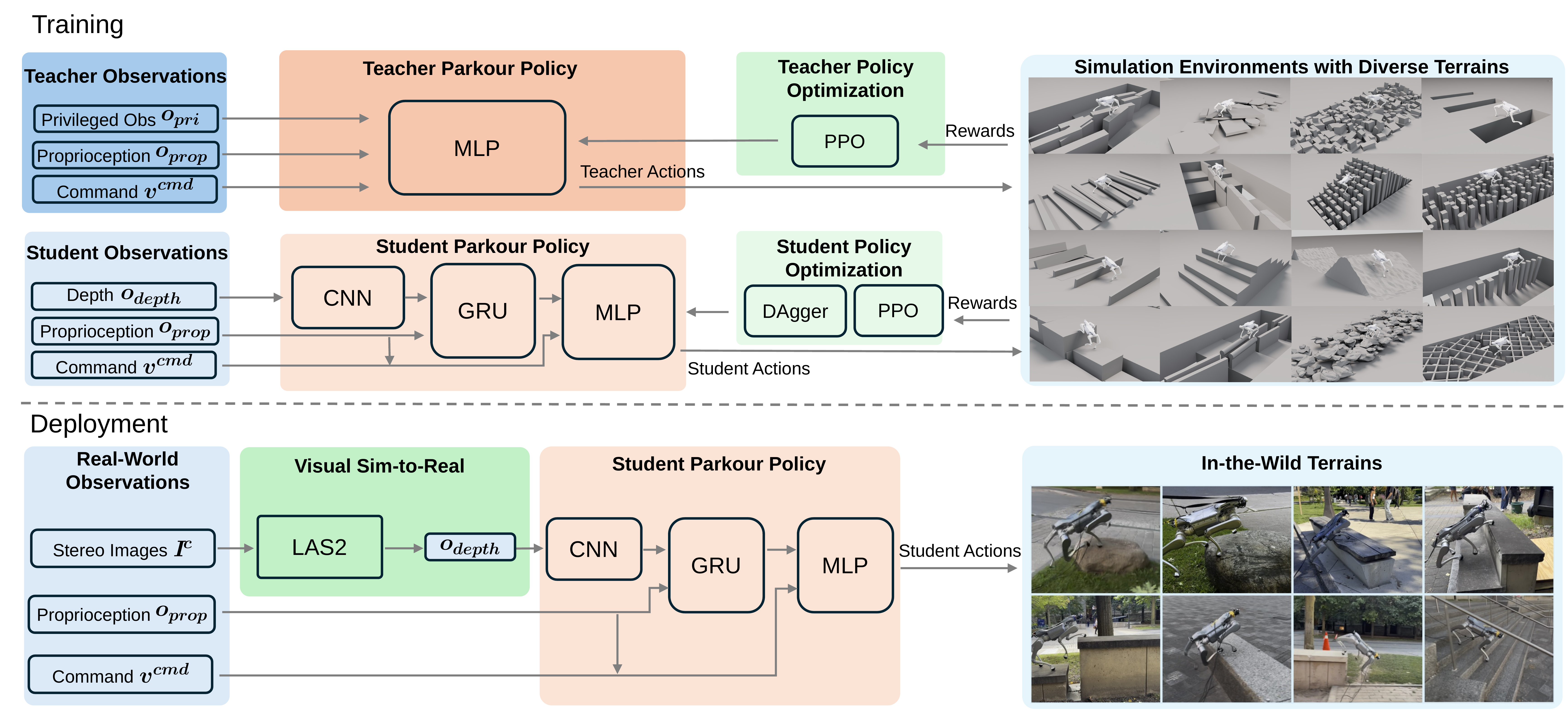}
\caption{The proposed ViSiR-Parkour architecture. Training: An MLP teacher policy maps privileged observations, proprioceptive observations, and velocity commands to actions and is trained using PPO on diverse simulated terrains. The student policy uses a CNN to encode depth images, a GRU to integrate visual and proprioceptive information across timesteps, and an MLP to generate actions conditioned on the GRU output, proprioception, and velocity commands. Hybrid policy learning combines DAgger for teacher action supervision with PPO for task reward optimization. Deployment: LAS2 estimates depth images from onboard stereo image pairs. These depth images, together with proprioceptive observations and velocity commands, enable the trained student policy to control the robot on unseen in-the-wild terrains.}
\label{fig:architecture}
\end{figure*}

The proposed ViSiR-Parkour architecture consists of training and deployment modules, Fig. \ref{fig:architecture}. These modules are discussed in detail below.

\subsection{Training}
During training, a single teacher policy is trained using DRL with privileged observations on diverse challenging terrains. An end-to-end visual student policy is then trained using a hybrid learning method that combines imitation learning and DRL.

\label{sec:method_training}

\subsubsection{Teacher Policy}\leavevmode\par\makeatletter\@afterheading\makeatother

\paragraph{Observations}
\label{sec:teacher_observations}
The teacher policy $\pi_\phi$ receives the observation $\boldsymbol{o}_{\phi,t}=(\boldsymbol{o}_{\mathrm{scan},t}^{b},\boldsymbol{o}_{\mathrm{scan},t}^{f},\boldsymbol{v}^{\mathrm{cmd}}_t,\boldsymbol{v}_t,\boldsymbol{o}_{\mathrm{prop},t})$ at each timestep $t$. The base linear velocity $\boldsymbol{v}_t$, $\boldsymbol{o}_{\mathrm{scan},t}^{b}$ and $\boldsymbol{o}_{\mathrm{scan},t}^{f}$ are privileged observations only available to the teacher policy. The base-centric multi-hit terrain scan $\boldsymbol{o}_{\mathrm{scan},t}^{b}=(\boldsymbol{o}_{\mathrm{scan},t}^{\mathrm{top}},\boldsymbol{o}_{\mathrm{scan},t}^{\mathrm{under}},\boldsymbol{o}_{\mathrm{scan},t}^{\mathrm{hidden}})\in\mathbb{R}^{3H_bW_b}$ represents the terrain surfaces within a local region around the robot base. Downward rays are cast over a grid. At each grid location, the multi-hit scan identifies: 1) the first surface hit by the downward rays, $\boldsymbol{o}_{\mathrm{scan},t}^{\mathrm{top}}\in\mathbb{R}^{H_bW_b}$; 2) the underside of that surface, $\boldsymbol{o}_{\mathrm{scan},t}^{\mathrm{under}}\in\mathbb{R}^{H_bW_b}$; and 3) the first upward-facing support surface below the underside, $\boldsymbol{o}_{\mathrm{scan},t}^{\mathrm{hidden}}\in\mathbb{R}^{H_bW_b}$. Together, these measurements describe terrain geometry for locomotion across diverse terrains, including the vertical clearance and support surfaces needed to traverse beneath overhead objects. The feet-centric height scan $\boldsymbol{o}_{\mathrm{scan},t}^{f}\in\mathbb{R}^{4N_f}$ represents the terrain height around the quadruped's four feet. Fig. \ref{fig:base_frame} shows the spatial distributions of the base-centric and feet-centric height-scan dots. In general, legged robot locomotion methods typically only use base-centric height scans to provide privileged observation of the local terrain geometry around the robot base \cite{parkourInTheWild}. We incorporate feet-centric height scans with a higher resolution than the base-centric height scans to provide three distinct benefits. Namely, the feet-centric height scans can provide: 1) the teacher policy with fine-grained terrain height information around each foot when traversing challenging discrete terrains such as stepping stones, as shown in Fig. \ref{fig:base_frame}; 2) a general terrain edge detector for reward shaping to penalize footholds that are too close to the terrain edges; and 3) foot-clearance measurements for reward shaping to encourage sufficient foot clearance and reduce foot dragging.

\paragraph{Scaling Terrain Diversity}
We scale the training terrain distribution along two dimensions: terrain-type diversity and intra-type variation. Terrain-type diversity represents the inclusion of structurally different terrains, such as slopes, stairs, gaps between separated support surfaces, and stepping stones. Intra-type variation consists of the geometric variation within one terrain type, such as stepping stone size, spacing between adjacent stepping stones, orientation of the stepping stones, and stone shape. Terrain-type diversity exposes the teacher policy to different foot-contact and locomotion requirements, while intra-type variation prevents the policy from associating a terrain type with a fixed geometry or layout. These variations are used together to train locomotion behaviors when the geometry or arrangement of a terrain differs from those encountered during training.

For terrain-type diversity, we divide the terrains into primitive terrains and combination terrains. A primitive terrain contains one terrain type throughout the course. We group primitive terrains according to their geometry for foot contact: 1) continuous-support terrains, such as slopes and rough surfaces, provide continuous surface support along the traversal path. The robot must adapt its gait and body posture to changes in the terrain surface \cite{challengingTerrainLocomotion}; 2) height-discontinuity terrains, such as stairs and raised platforms, contain abrupt height changes along the traversal path. The robot must step, climb, or jump across these height changes \cite{anymalParkour,robotParkourLearning,extremeParkour}; and 3) sparse-support terrains, such as stepping stones and balance beams, provide narrow or separated surface support along the traversal path. The robot must place its feet and maintain balance on the available support regions \cite{START,walkThePLANC}.

A combination terrain contains a sequence of terrain segments generated from different primitive terrain types. We randomly sample the geometry of each segment and the longitudinal gap, lateral offset, and height difference between adjacent segments. By varying the boundary between segments, combination terrains require the teacher policy to acquire cross-terrain transition skills that are not exercised on isolated primitive terrains. For example, transitioning from stepping stones directly onto a high platform requires precise foothold placement and agile climbing in a single maneuver, a compound skill not acquired by traversing either terrain alone.

The intra-type variation is increased by resampling two groups of terrain-specific parameters for each terrain instance: 1) geometric parameters, which vary terrain dimensions and relative placements, such as obstacle height and foothold spacing; and 2) structural parameters, which vary discrete forms and layouts, such as straight or curved stairs and flat or rounded footholds. A normalized difficulty variable controls the geometric parameters that directly affect traversal difficulty, while the remaining geometric and structural parameters are sampled independently within their terrain-specific ranges. This intra-type variation prevents the policy from overfitting to fixed terrain configurations and supports generalization to unseen geometries and layouts.

\paragraph{PATR Curriculum Learning}
Our PATR curriculum technique adjusts each robot's difficulty level and reallocates robots across terrain types according to terrain traversal performance. We train the teacher policy using $N_{\phi,\mathscr{r}}$ simulated robots in parallel \cite{learningToWalkInMinutes}, with each robot assigned to one terrain instance. We use $L=10$ difficulty levels, indexed from 0 to $L-1$, for each terrain type, with higher levels using more challenging terrain parameters. At the start of training, each robot $\mathscr{r}_i$ is assigned a fixed home terrain type $h_i$ and starts at difficulty level 0. For difficulty-level progression, we follow the game-inspired curriculum in \cite{learningToWalkInMinutes}, advancing robots to more difficult levels within their home terrain types after successful traversal. Our curriculum additionally reallocates robots from terrain types with high traversal success rates to those with lower success rates, directing extra training toward terrain types the policy has not yet learned to traverse reliably.

\suppressfloats[t]
\begingroup
\makeatletter
\renewcommand\fs@ruled{%
  \def\@fs@cfont{\bfseries\color{black}}%
  \def\@fs@capt##1##2{{\color{black}\floatc@ruled{##1}{##2}}}%
  \def\@fs@pre{{\color{black}\hrule height.8pt depth0pt}\kern2pt}%
  \def\@fs@post{\kern2pt{\color{black}\hrule}\relax}%
  \def\@fs@mid{\kern2pt{\color{black}\hrule}\kern2pt}%
  \let\@fs@iftopcapt\iftrue}
\makeatother
\begin{algorithm}[t]
\color{black}
\caption{Progress-aware terrain reallocation curriculum.}
\label{alg:terrain_curriculum}
\footnotesize
\begin{algorithmic}[1]
\REQUIRE Terrain types $\mathcal{K}$, levels $L$, terrain instance sets $\mathcal{T}_{k,j}$ (type $k$, level $j$), $d_{\mathrm{up}}$, $\rho_{\mathrm{down}}$, $s_{\mathrm{cut}}$, and $\alpha$
\STATE Assign each robot $\mathscr{r}_i$ a fixed home terrain type $h_i\in\mathcal{K}$
\STATE Set stored home level $\ell_i\gets0$ for every robot $\mathscr{r}_i$
\STATE Initialize traversal outcome records for each terrain instance and level
\WHILE{training is in progress}
    \STATE $\mathcal{I}\gets\{i\mid\text{robot }\mathscr{r}_i\text{ has completed a parkour episode}\}$
    \STATE Reduce the weight of older episode outcomes
    \FOR{each robot $\mathscr{r}_i$ with $i\in\mathcal{I}$}
        \STATE Compute $d_i$ and $\rho_i$ using (\ref{eq:curriculum_progress})
        \STATE Record success ($d_i>d_{\mathrm{up}}$) or failure on the completed terrain instance and level
        \IF{the completed episode was on $h_i$}
            \IF{$d_i>d_{\mathrm{up}}$}
                \IF{$\ell_i=L-1$}
                    \STATE $\ell_i\sim\mathcal{U}\{0,\ldots,L-1\}$
                \ELSE
                    \STATE $\ell_i\gets\ell_i+1$
                \ENDIF
            \ELSIF{$\rho_i<\rho_{\mathrm{down}}$}
                \STATE $\ell_i\gets\max(\ell_i-1,0)$
            \ENDIF
        \ENDIF
    \ENDFOR
    \FOR{each terrain type $k\in\mathcal{K}$ and level $j\in\{0,\ldots,L-1\}$}
        \STATE $s_{k,j}\gets$ mean recent success rate across terrain instances with sufficient episode data
    \ENDFOR
    \STATE Compute $s_k$ and $g_k$ using (\ref{eq:terrain_priority})
    \STATE $\bar{\ell}_k\gets\frac{\sum_{i:h_i=k}\ell_i}{|\{i:h_i=k\}|}$ for all $k\in\mathcal{K}$
    \STATE $p_{\mathrm{realloc}}\gets(1-\alpha)\min\left(1,\frac{\sum_{k\in\mathcal{K}}g_k}{s_{\mathrm{cut}}}\right)$
    \FOR{each robot $\mathscr{r}_i$ with $i\in\mathcal{I}$}
        \STATE $u_i\sim\mathcal{U}(0,1)$
        \IF{$s_{h_i}\geq s_{\mathrm{cut}}$ \AND $u_i<p_{\mathrm{realloc}}$}
            \STATE $p^{\mathrm{dest}}_{ik}\gets\frac{\mathbf{1}[k\neq h_i]g_k^2}{\sum_{m\in\mathcal{K}\setminus\{h_i\}}g_m^2}$ for all $k\in\mathcal{K}$
            \STATE $\boldsymbol{p}^{\mathrm{dest}}_i\gets(p^{\mathrm{dest}}_{ik})_{k\in\mathcal{K}}$
            \STATE $k_i\sim\operatorname{Categorical}(\boldsymbol{p}^{\mathrm{dest}}_i)$
            \STATE $\ell_i^{\mathrm{dest}}\gets\mathrm{round}(\bar{\ell}_{k_i})$
            \STATE $\xi_i\sim\mathcal{U}(\mathcal{T}_{k_i,\ell_i^{\mathrm{dest}}})$
            \STATE Place robot $\mathscr{r}_i$ on terrain instance $\xi_i$
        \ELSE
            \STATE Place robot $\mathscr{r}_i$ on its home terrain instance at level $\ell_i$
        \ENDIF
    \ENDFOR
\ENDWHILE
\end{algorithmic}
\end{algorithm}
\endgroup

To update each robot's difficulty level $\ell_i$ within its home terrain type $h_i$, we evaluate terrain completion progress after each episode. We compute the forward distance $d_i$ and the commanded-distance ratio $\rho_i$:
\begin{equation}
d_i=\max\left(p^x_{i,T}-p^x_{i,0},0\right), \qquad
\rho_i=\frac{d_i}{\lVert\boldsymbol{v}^{\mathrm{cmd}}_{xy,i}\rVert_2T_e},
\label{eq:curriculum_progress}
\end{equation}
\noindent where $p^x_{i,0}$ and $p^x_{i,T}$ are the robot's starting position and the robot's final position along the traversal direction, and $T_e$ is the episode time limit. When $d_i>d_{\mathrm{up}}$, the traversal is considered successful and the home level $\ell_i$ increases by one. $\ell_i$ decreases by one when $\rho_i<\rho_{\mathrm{down}}$ and otherwise remains unchanged. $d_{\mathrm{up}}$ is the forward-distance threshold for advancing to the next difficulty level and $\rho_{\mathrm{down}}$ is the commanded-distance-ratio threshold for decreasing the difficulty level. After successfully traversing the highest difficulty level, the robot's home difficulty level is reset to a random level.

\begingroup
\setlength{\parskip}{0pt}
We evaluate the policy's traversal performance on each terrain type using a success score. For terrain type $k$, $s_{k,j}$ is the mean traversal success rate across its instances at difficulty level $j$. We average these rates across difficulty levels to obtain the terrain-type success score $s_k$ and compute its training-priority score $g_k$:
\begin{equation}
s_k=\frac{1}{L}\sum_{j=0}^{L-1}s_{k,j},\qquad g_k=\max(0,s_{\mathrm{cut}}-s_k),
\label{eq:terrain_priority}
\end{equation}
\endgroup
\noindent where $s_{\mathrm{cut}}$ is the success threshold, whose value is determined empirically based on the robot's ability to traverse different terrain types during training. Lower success scores give higher training priority. Only robots whose home terrain types meet this threshold are eligible for reallocation. At each episode reset, each eligible robot is reallocated with probability:
\begin{equation}
p_{\mathrm{realloc}}=(1-\alpha)\min\left(1,\frac{\sum_{k\in\mathcal{K}}g_k}{s_{\mathrm{cut}}}\right),
\label{eq:terrain_reallocation_probability}
\end{equation}
\noindent where $\mathcal{K}$ is the set of terrain types and $1-\alpha$ is the maximum reallocation probability. For a robot $\mathscr{r}_i$ selected for reallocation, we sample a destination terrain type $k$ with probability $p^{\mathrm{dest}}_{ik}$:
\begin{equation}
p^{\mathrm{dest}}_{ik}=\frac{g_k^2}{\sum_{m\in\mathcal{K}\setminus\{h_i\}}g_m^2}\text{.}
\label{eq:terrain_destination_probability}
\end{equation}
Robots that are not reallocated return to their home terrain types at their home difficulty levels. When all terrain types meet the success threshold, all priority scores are zero and no robots are reallocated. Alg. 1 provides the overall procedure.

\paragraph{Policy Optimization}
We optimize the teacher policy over the resulting curriculum using DRL. The teacher policy $\pi_\phi$ is implemented as a multilayer perceptron (MLP) trained using Proximal Policy Optimization (PPO) \cite{PPO}. PPO is used due to its stability and effectiveness for legged robot locomotion \cite{learningToWalkInMinutes}. An asymmetric actor--critic architecture \cite{asymmetricActorCritic} was utilized, where the actor receives noisy simulated observations and the critic receives the corresponding clean simulated observations. The teacher actor and critic are optimized using the PPO loss with $\psi_\phi$ denoting the teacher critic parameters:
\begin{equation}
\mathcal{L}_{\mathrm{PPO}}(\phi,\psi_\phi)
=\mathcal{L}_{\mathrm{clip}}(\phi)
+c_v\mathcal{L}_{\mathrm{value}}(\psi_\phi)
-c_e\mathcal{H}(\pi_\phi),
\label{eq:ppo_loss}
\end{equation}
\noindent where $\mathcal{L}_{\mathrm{clip}}$ is the clipped surrogate loss, $\mathcal{L}_{\mathrm{value}}$ is the critic value loss, $\mathcal{H}$ is the policy entropy, and $c_v$ and $c_e$ are the value-loss and entropy coefficients, respectively. We apply left-right symmetry augmentation \cite{symmetryRobotPolicies} to encourage symmetric gait.

During training, the episode termination conditions include episode timeout, the robot base contacting the terrain or an obstacle, the robot falling into gaps, and excessive changes in commanded joint actions ($\|\boldsymbol{a}^{\phi}_t-\boldsymbol{a}^{\phi}_{t-1}\|^2>\delta_a$), where $\delta_a$ is the action-change termination threshold, determined empirically by evaluating the current action with respect to the previous action during training. Large action changes can make the training unstable. Robot base contact with the terrain or an obstacle as a termination condition is typical for legged robot locomotion \cite{AME}. However, immediate episode termination upon base contact with the terrain or an obstacle can create a safety risk during real-world deployment. If the robot fails (e.g. falls) and its base contacts the terrain, the policy enters an out-of-distribution state that it has not experienced during training, generating unpredictable actions. Our policy is exposed to failure states during training by delaying termination until the base contact continues for 0.1 s (five control steps), allowing it to learn safe actions before episode termination. We selected the 0.1 s duration empirically by comparing short and long termination windows during training: a long window can make the policy too conservative and unwilling to traverse challenging terrains, whereas a short window provides too few training samples.

\begin{table}[t]
\caption{Simulation Parameters}
\label{tab:simulation_parameters}
\centering
\scriptsize
\setlength{\tabcolsep}{1.5pt}
\renewcommand{\arraystretch}{0.95}
\begin{tabular}{@{}p{0.21\columnwidth}p{0.25\columnwidth}p{0.21\columnwidth}p{0.25\columnwidth}@{}}
\toprule
\multicolumn{1}{c}{Parameter} & \multicolumn{1}{c}{Value} & \multicolumn{1}{c}{Parameter} & \multicolumn{1}{c}{Value} \\
\midrule
Simulation step & 0.005~s & Policy frequency & 50~Hz \\
$T_e$ & 15~s & $\boldsymbol{s}_a$ & 0.25 \\
$\boldsymbol{K}_p$ & 118.4~N$\cdot$m/rad & $\boldsymbol{K}_d$ & 7.54~N$\cdot$m$\cdot$s/rad \\
$H_b$ & 21 & $W_b$ & 11 \\
$N_f$ & 5 & Height-scan noise & $[-0.02,0.02]$~m \\
$\boldsymbol{o}_{\mathrm{scan}}^b$ resolution & 0.1~m & $\boldsymbol{o}_{\mathrm{scan}}^f$ resolution & 0.05~m \\
$\boldsymbol{o}_{\mathrm{depth}}$ & $2{\times}48{\times}64$ & Depth range / rate & 2~m / 16.7~Hz \\
$v_x^{\mathrm{cmd}}$ & $[0.3,1.2]$~m/s & $v_{\mathrm{th}}$ & 0.5~m/s \\
$\omega_z^{\mathrm{cmd}}$ & $[-1.2,1.2]$~rad/s & $\omega_{\mathrm{th}}$ & 0.2~rad/s \\
$v_y^{\mathrm{cmd}}$ & 0~m/s & Heading range & $[-\pi,\pi]$~rad \\
\midrule
Friction & $[0.3,1.2]$ & Restitution & $[0,0.15]$ \\
Base mass offset & $[-3,5]$~kg & $\boldsymbol{q}^{\mathrm{nom}}$ offset & $[-0.01,0.01]$~rad \\
$T_{\mathrm{dist}}$ & $[4,6]$~s & Camera position & $[-2,2]$~cm \\
Push $\boldsymbol{v}_{xy}$ & $[-0.5,0.5]$~m/s & Push $\boldsymbol{\omega}_{xy}$ & $[-0.52,0.52]$~rad/s \\
Push $v_z$ & $[-0.2,0.2]$~m/s & Push $\omega_z$ & $[-0.78,0.78]$~rad/s \\
Camera rotation & $[-3,3]^\circ$ & $\boldsymbol{v}_{xy},v_z$ noise & $[-0.1,0.1]$~m/s \\
$\boldsymbol{\omega}$ noise & $[-0.2,0.2]$~rad/s & $\boldsymbol{g}$ noise & $[-0.05,0.05]$ \\
$\boldsymbol{q}$ noise & $[-0.01,0.01]$~rad & $\dot{\boldsymbol{q}}$ noise & $[-0.5,0.5]$~rad/s \\
\bottomrule
\end{tabular}
\end{table}

\begin{table}[t]
\caption{Rewards}
\label{tab:rewards}
\centering
\scriptsize
\setlength{\tabcolsep}{1pt}
\begin{tabular}{clll}
\toprule
Type & Description & Equation & Weight \\
\midrule
\multirow{3}{*}{\rotatebox[origin=c]{90}{Task}}
& Linear velocity tracking & $\exp(-\|\boldsymbol{v}_{xy}^{\mathrm{cmd}}-\boldsymbol{v}_{xy}\|^2/0.25)$ & $1.5$ \\
& Angular velocity tracking & $\exp(-(\omega_z^{\mathrm{cmd}}-\omega_z)^2/0.25)$ & $0.5$ \\
& Forward progress & $\mathbf{1}^{\mathrm{move}}[d_t-\max_{t'<t}d_{t'}]_+$ & $20$ \\
\midrule
\multirow{18}{*}{\rotatebox[origin=c]{90}{Regularization}}
& Vertical velocity & $v_z^2$ & $-2$ \\
& Roll--pitch velocity & $\|\boldsymbol{\omega}_{xy}\|^2$ & $-0.05$ \\
& Joint torques & $\|\boldsymbol{\tau}\|^2$ & $-2{\times}10^{-5}$ \\
& Joint accelerations & $\|\ddot{\boldsymbol{q}}\|^2$ & $-2.5{\times}10^{-7}$ \\
& Mechanical energy & $\sum_j|\tau_j\dot{q}_j|$ & $-2{\times}10^{-4}$ \\
& Action rate & $\|\boldsymbol{a}_t-\boldsymbol{a}_{t-1}\|^2$ & $-0.05$ \\
& Action smoothness & $\|\boldsymbol{a}_t-2\boldsymbol{a}_{t-1}+\boldsymbol{a}_{t-2}\|^2$ & $-0.01$ \\
& Joint position limits & $(1{+}\mathbf{1}^{\mathrm{fall}})\|[\boldsymbol{q}{-}\overline{\boldsymbol{q}}^{\mathrm{lim}}]_+{+}[\underline{\boldsymbol{q}}^{\mathrm{lim}}{-}\boldsymbol{q}]_+\|_1$ & $-300$ \\
& Joint velocity limits & $\|[\,|\dot{\boldsymbol{q}}|-0.9\,\dot{\boldsymbol{q}}^{\mathrm{lim}}]_+\|_1$ & $-10$ \\
& \multirow{3}{*}{Reference joint deviation} & $\mathbf{1}^{\mathrm{stand}}\|\boldsymbol{q}-\boldsymbol{q}^{\mathrm{ref}}\|^2$ & $-3.75$ \\
& & $\mathbf{1}^{\mathrm{move}}\sum_{j\in\mathcal{J}_{\mathrm{hip}}}(q_j-q_j^{\mathrm{ref}})^2$ & $-0.75$ \\
& & $\mathbf{1}^{\mathrm{move}}\sum_{j\in\mathcal{J}_{\mathrm{thigh,calf}}}(q_j-q_j^{\mathrm{ref}})^2$ & $-0.2$ \\
& Feet air time & $\mathbf{1}^{\mathrm{move}}\sum_f(t_f^{\mathrm{air}}-0.5)\,\mathbf{1}_f^{\mathrm{land}}$ & $0.05$ \\
& Feet slide & $\sum_f\|\boldsymbol{v}_{f,xy}\|\,\mathbf{1}_f^{\mathrm{contact}}$ & $-0.1$ \\
& Feet stumble & $\mathbf{1}[\exists f{:}\,\|\boldsymbol{F}_{f,xy}\|>4|F_{f,z}|]$ & $-1$ \\
& Feet clearance & $\exp(-10^{3}\sum_f[0.08{-}h_f]_+^2\|\boldsymbol{v}_{f,xy}\|)$ & $0.15$ \\
& Feet edge landing & $\sum_f\mathbf{1}_f^{\mathrm{land}}\mathbf{1}[\max_k|h_{f,k}{-}h_{f,0}|{>}0.1]$ & $-5$ \\
& Undesired contact & $(1{-}\mathbf{1}^{\mathrm{fall}})\sum_{b\in\mathcal{B}}\mathbf{1}[\|\boldsymbol{F}_b\|{>}1]$ & $-1$ \\
\bottomrule
\end{tabular}
\end{table}

For dynamics sim-to-real transfer, we construct the simulated robot using the Unified Robot Description Format (URDF), which provides the measured robot geometry, inertial properties, joint limits, and motor torque limits. We set the motor PD gains following the heuristics in \cite{beyondMimic}. We apply DR to improve the robustness of the teacher policy by exposing it to variations in robot and environment dynamics during training. The robot-terrain friction, base mass, and default joint positions are uniformly sampled from the ranges specified in Table \ref{tab:simulation_parameters} during simulation initialization. Furthermore, external push disturbances are simulated as instantaneous changes to the robot base velocity. At intervals $T_{\mathrm{dist}}$, randomly sampled linear and angular velocity perturbations are applied to the robot base; the sampling ranges for the disturbance interval and velocity perturbations are provided in Table \ref{tab:simulation_parameters}.

\paragraph{Rewards}
The reward function, $r$, consists of task rewards and regularization rewards. The task rewards encourage the robot to track a given velocity command, while making forward progress along the desired direction. The regularization rewards shape the robot behavior so that it moves smoothly, naturally, and efficiently. The detailed reward definitions and their corresponding weights are presented in Table \ref{tab:rewards}. The feet edge landing and feet clearance rewards use the feet-centric height scan introduced in Section IV-A to penalize footholds close to terrain edges and to encourage the swing feet to remain clear of the terrain during foot motion, thereby reducing dragging. While the robot base is in contact with the terrain during the termination window ($\mathbf{1}^{\mathrm{fall}}{=}1$), the undesired contact penalty is set to zero and the joint position limit penalty is increased by a factor of 2. This prevents the policy from becoming over conservative due to the large contact penalties accumulated during a fall, while also discouraging dangerous joint motions that can damage the robot's motors.

\subsubsection{Student Policy}
During training, depth images are generated by casting rays from each simulated camera onto the terrain and visible robot meshes. The resulting simulated depth images are clean and provide exact ray-to-surface distances for the simulated scene, including self-occlusions from the robot legs.

\paragraph{Policy Network}
The student policy $\pi_\theta$ consists of a CNN, a gated recurrent unit (GRU) \cite{gatedRecurrentUnit}, and an MLP, Fig. \ref{fig:architecture}. At each timestep, the two depth images $\boldsymbol{o}_{\mathrm{depth},t}$ (forward-facing and downward-facing) are stacked as separate channels and encoded by the CNN into a visual feature $\boldsymbol{z}^{\mathrm{vis}}_t$. This feature is concatenated with the proprioceptive observation $\boldsymbol{o}_{\mathrm{prop},t}$ and mapped by a fusion layer to $\boldsymbol{z}^{\mathrm{fuse}}_t$. The GRU combines $\boldsymbol{z}^{\mathrm{fuse}}_t$ with its previous hidden state $\boldsymbol{h}_{t-1}$ to output the updated hidden state $\boldsymbol{h}_t$, which retains terrain and robot state information across timesteps. Finally, the MLP maps the concatenation of the hidden state $\boldsymbol{h}_t$, the current proprioceptive observation $\boldsymbol{o}_{\mathrm{prop},t}$, and the velocity command $\boldsymbol{v}^{\mathrm{cmd}}_t$ to the action $\boldsymbol{a}_t$.

\paragraph{Hybrid Policy Learning}
We train the student policy network using a hybrid policy learning method that combines imitation learning and DRL to transfer the teacher's locomotion behavior while allowing the student to optimize task performance without being restricted to reproducing the teacher's actions \cite{AME2,hamlet}. DAgger \cite{DAgger} is used for imitation learning to provide teacher action supervision at states visited by the student, helping it correct errors after deviating from the teacher's actions. PPO \cite{PPO} is used as the DRL method to optimize the task rewards, allowing the student to improve task performance in addition to learning to match the teacher's actions. PPO uses the visual student policy $\pi_\theta$ as the actor and an MLP critic with parameters $\psi_\theta$ that receives clean privileged observations during training. During training, the actions generated by the student policy are executed in simulation. At each timestep, the frozen teacher policy receives the privileged observation $\boldsymbol{o}_{\phi,t}$ for the same simulated robot and generates the target action $\boldsymbol{a}^{\phi}_t=\pi_\phi(\boldsymbol{o}_{\phi,t})$. The student observation and corresponding teacher action are stored in $\mathcal{D}=\{(\boldsymbol{o}_t,\boldsymbol{a}^{\phi}_t)\}$. These paired student observations and teacher actions form the supervision for the DAgger imitation loss \cite{DAgger}, defined as the negative log-likelihood of the teacher action under the student policy:
\begin{equation}
\mathcal{L}_{\mathrm{DAgger}}(\theta)=-\frac{1}{|\mathcal{D}|}
\sum_{(\boldsymbol{o}_t,\boldsymbol{a}^{\phi}_t)\in\mathcal{D}}
\log \pi_\theta(\boldsymbol{a}^{\phi}_t\mid\boldsymbol{o}_t).
\end{equation}
\noindent We combine the DAgger imitation loss with the PPO loss to optimize the student policy:
\begin{equation}
\mathcal{L}_{\mathrm{hybrid}}(\theta,\psi_\theta)
=\mathcal{L}_{\mathrm{PPO}}(\theta,\psi_\theta)
+\beta\mathcal{L}_{\mathrm{DAgger}}(\theta),
\end{equation}
\noindent where $\beta$ is the DAgger loss coefficient.

\begin{figure}[!t]
\centering
\includegraphics[width=0.85\columnwidth]{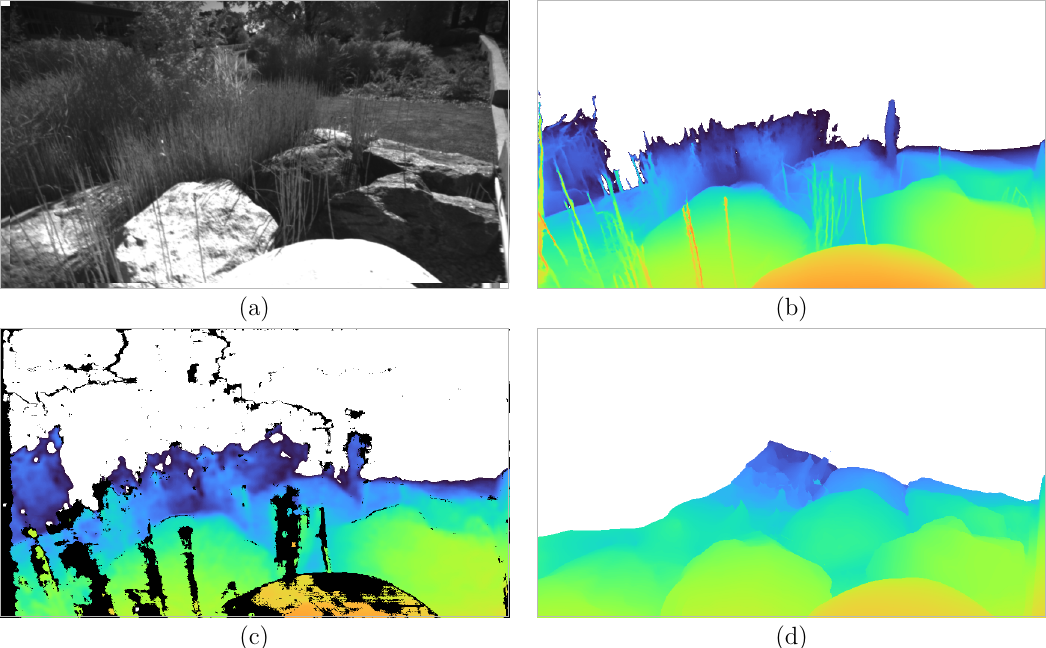}
\caption{Visual sim-to-real depth comparison on outdoor rock terrain: (a) raw stereo image pair, (b) LAS2 depth image output, (c) raw onboard depth image, and (d) simulated depth from 3D scanned real-world rock terrain.}
\label{fig:stereo_matching}
\end{figure}

Student training begins with a warm-up phase in which the DAgger loss trains the actor while the value loss trains the critic. We use the warm-up phase to train the critic before PPO updates the actor as inaccurate predictions of future rewards from an untrained critic can destabilize policy training \cite{parkourInTheWild,hamlet}. After the warm-up, we enable the PPO clipped surrogate and entropy terms while keeping the DAgger loss active, so that task rewards and teacher supervision jointly optimize the student. We do not apply left-right symmetry augmentation \cite{symmetryRobotPolicies} during student training as learned hidden states of the GRU cannot be mirrored by explicitly exchanging left- and right-leg joint variables, as done for robot observations and actions. Instead, we encourage symmetric gait through teacher action supervision. We linearly decrease the entropy coefficient so that PPO explores more when first enabled and reduces action randomness later in training. We retain the base-contact termination delay during hybrid policy learning, allowing the student to imitate the teacher's safer actions after base contact with the terrain or an obstacle to reduce the risk of damaging joint motors during real-world deployment.

\subsection{Deployment}
\label{sec:method_deployment}
During deployment, a stereo matching model estimates depth images from the onboard stereo observations to reduce the visual sim-to-real gap. Unlike the clean simulated depth images used during training, raw depth images from onboard stereo cameras on the physical robot are noisy and can contain missing pixels and inaccurate depth estimates \cite{nowYouSeeThat,hikingInTheWild}, creating a visual sim-to-real gap for the student policy.

\nocite{isaacLab} %
\suppressfloats[t]
\begin{table}[t]
\caption{Policy Training Hyperparameters}
\label{tab:training_hyperparameters}
\centering
\footnotesize
\setlength{\tabcolsep}{3pt}
\begin{tabular}{@{}lcc@{}}
\toprule
Parameter & Teacher & Student \\
\midrule
Training method & PPO & PPO + DAgger \\
Parallel robots & 8,192 & 4,096 \\
Training iterations & 20,000 & 20,000 \\
Steps per iteration & 24 & 24 \\
Optimization epochs & 5 & 5 \\
Optimizer & Adam~\cite{adam} & Adam~\cite{adam} \\
Learning rate & $1\times10^{-3}$ & $1\times10^{-3}$ \\
Maximum gradient norm & 1.0 & 1.0 \\
Curriculum success threshold $s_{\mathrm{cut}}$ & 0.8 & 0.8 \\
Reallocation $\alpha$ & 0.4 & 0.4 \\
Action-change threshold $\delta_a$ & $10^{4}$ & $10^{4}$ \\
\midrule
Actor architecture & MLP & CNN, GRU, MLP \\
CNN channels & -- & 24, 48, 64, 64 \\
GRU dimension & -- & 256 \\
GRU sequence length & -- & 24 \\
MLP dimensions & 512, 512, 256, 128 & 128 \\
PPO mini-batches & 4 & 4 \\
PPO clip parameter & 0.2 & 0.2 \\
PPO $\gamma$ & 0.99 & 0.99 \\
GAE $\lambda$ & 0.95 & 0.95 \\
Value loss $c_v$ & 1.0 & 1.0 \\
Entropy $c_e$ & $0.01\rightarrow0.001$ & $0.004\rightarrow0.001$ \\
Desired KL & 0.01 & 0.01 \\
DAgger $\beta$ & -- & 0.02 \\
Warm-up iterations & -- & 2,000 \\
\bottomrule
\end{tabular}
\end{table}

To reduce this gap, the visual sim-to-real module uses the learning-based stereo matching model Lite Any Stereo V2 (LAS2) \cite{LAS2} for its strong zero-shot generalization and real-time inference speed. For each camera $c\in\{\mathrm{front},\mathrm{down}\}$ at timestep $t$, the model takes the rectified stereo image pair $\boldsymbol{I}_t^c=(\boldsymbol{I}_{L,t}^c,\boldsymbol{I}_{R,t}^c)$ as input, where $\boldsymbol{I}_{L,t}^c$ and $\boldsymbol{I}_{R,t}^c$ denote the left and right images, respectively. The stereo matching model uses shared-weight CNN encoders to extract features from the stereo image pair, and applies 2D cost aggregation to estimate a disparity image. We convert the disparity to metric depth using the calibrated camera focal length and stereo baseline, and resize and clip the depth images to the resolution and range used during training. Fig. \ref{fig:stereo_matching} shows an example of the raw stereo pair, LAS2 depth estimate, raw onboard depth, and simulated depth. Compared with the raw camera depth, the LAS2 depth images contain less noise and provide more complete and accurate terrain geometry, allowing the student policy trained on clean simulated depth images to transfer without depth-noise augmentation.

Using the depth observations generated by the visual sim-to-real module during deployment, the student policy $\pi_\theta$ generates actions $\boldsymbol{a}_t$ for execution. As the deployed student policy receives the same depth representation used during simulation training through the visual sim-to-real module, the policy can be deployed without real-world fine-tuning.

\section{Training and Deployment Set-up}
\subsection{Training Set-up}
Both the teacher and student policies are trained in Isaac Lab \cite{isaacLab} using a simulated Unitree A2 quadruped robot on an NVIDIA GeForce RTX 5090 GPU workstation. We incorporate 37 primitive terrain types and combination terrains with up to 1,332 directed transitions between distinct primitive terrains ($\frac{37!}{(37-2)!}$). For each episode, a candidate forward velocity $\tilde{v}_x\sim\mathcal{U}(v_{x,\min}^{\mathrm{cmd}},v_{x,\max}^{\mathrm{cmd}})$ is sampled. If $\tilde{v}_x\geq v_{\mathrm{th}}$, we set $v_x^{\mathrm{cmd}}=\tilde{v}_x$ and $v_y^{\mathrm{cmd}}=0$ for forward parkour. Otherwise, we set $\boldsymbol{v}_{xy}^{\mathrm{cmd}}=0$ and sample $\tilde{\omega}_z\sim\mathcal{U}(\omega_{z,\min}^{\mathrm{cmd}},\omega_{z,\max}^{\mathrm{cmd}})$. We set $\omega_z^{\mathrm{cmd}}=\tilde{\omega}_z$ when $|\tilde{\omega}_z|\geq\omega_{\mathrm{th}}$ for in-place rotation and $\omega_z^{\mathrm{cmd}}=0$ otherwise for the robot standing still. This command sampling scheme trains a single policy to perform parkour, rotations, and standing actions. The simulation and policy training parameters are provided in Tables \ref{tab:simulation_parameters} and \ref{tab:training_hyperparameters}, respectively. The values of the hyperparameters were empirically determined during training.

\subsection{Deployment Set-up}
The student policy is deployed on the physical A2 robot with onboard forward-facing and downward-facing Intel RealSense D435i stereo cameras. LAS2-L \cite{LAS2} is used for stereo depth estimation, due to its high stereo-matching accuracy and real-time inference on our deployment hardware. LAS2-L processes each stereo image pair at the native 848$\times$480 resolution, and the depth output is downsampled to 48$\times$64 before being provided to the student policy. LAS2-L and the student parkour policy run on a laptop with an Intel Core i7-12700H CPU and an NVIDIA GeForce RTX 3050 Ti GPU. Both LAS2-L and the student policy are exported to ONNX \cite{ONNX}; LAS2-L is converted to a TensorRT engine \cite{TensorRT} for GPU inference, while the student policy is executed using ONNX Runtime on the CPU.

\begin{table*}[!b]
\caption{Comparison of ViSiR-Parkour with SOTA Visual Parkour Methods}
\label{tab:simulated_comparison}
\centering
\scriptsize
\setlength{\tabcolsep}{3pt}
\renewcommand{\arraystretch}{1.1}
\begin{tabular*}{\textwidth}{@{\extracolsep{\fill}}lccccccc@{}}
\toprule
\multirow{2}{*}{Method} & \multicolumn{3}{c}{In-Distribution Terrains, SR (\%)} & \multicolumn{3}{c}{Out-of-Distribution Terrains, SR (\%)} & \multirow{2}{*}{Overall SR (\%)} \\
\cmidrule(lr){2-4}\cmidrule(lr){5-7}
& Primitive & Combination & Average & Unseen Primitive & Unseen Combination & Average & \\
\midrule
EP & 58.2& 48.5 & 53.4 & 30.3 & 12.1 & 21.2 & 37.3 \\
START & 62.5 & 47.2 & 54.8 & 26.6 & 11.5 & 19.1 & 37.0 \\
PitW & 70.3 & 59.4 & 64.8 & 31.1 & 20.0 & 25.6 & 45.2 \\
\shortstack{\textbf{ViSiR-Parkour} } & \textbf{92.4} & \textbf{82.2} & \textbf{87.3} & \textbf{66.9} & \textbf{38.3} & \textbf{52.6} & \textbf{69.9} \\
\bottomrule
\end{tabular*}
\end{table*}

\begin{figure}[!t]
\centering
\includegraphics[width=\columnwidth]{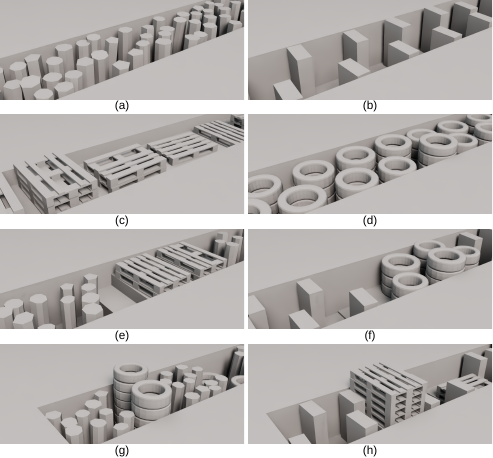}
\caption{Unseen primitive terrains: (a) irregular stepping stones, (b) staggered short beams, (c) stacked pallets, and (d) stacked rings. Unseen combination terrains with gaps: (e) stones and pallets, (f) beams and rings; and combination terrains with height changes: (g) stones and rings, (h) beams and pallets.}
\label{fig:unseen_terrains}
\end{figure}

\section{Simulated Experiments}
ViSiR-Parkour was evaluated by conducting the following simulated experiments: 1) a comparison study with the state-of-the-art (SOTA) visual parkour methods to assess its performance and generalizability on challenging terrains, 2) a scalability study to evaluate ViSiR-Parkour's generalizability to out-of-distribution terrains when trained with an increasing number of terrain types, and 3) ablation studies to analyze the impact of the major design components of ViSiR-Parkour.
\subsection{Comparison Study with SOTA Visual Parkour Methods}
ViSiR-Parkour was compared with SOTA visual parkour methods with respect to its locomotion performance on in-distribution terrains and its generalizability to out-of-distribution primitive terrain types and configurations.

The in-distribution terrains consist of terrain types and configurations represented during training. They include the complete set of primitive terrains and the combination terrains formed by sequences of different primitive terrains. The out-of-distribution terrains consist of both unseen primitive and combination terrains. The four unseen primitive terrain types are, Fig. \ref{fig:unseen_terrains} (a)-(d): 1) irregular stepping stones, with nonuniform polygonal shapes and placements; 2) staggered short beams, which extend from alternating sides of the terrain; 3) stacked pallets, with gaps between the boards forming their top surfaces; and 4) stacked rings, consisting of rigid rings with rounded upper surfaces and central openings. These terrains were selected to evaluate whether the parkour policy can adapt the robot's foothold placement to unseen foothold geometries and support layouts. The unseen combination terrains combine these four unseen primitive terrain types in pairs, with gaps or height changes between consecutive terrain segments, Fig. \ref{fig:unseen_terrains}(e)-(h). These terrains evaluate whether the parkour policy can generalize to transitions between unseen terrain types. The performance metric used is success rate (SR), defined as the percentage of trials in which the robot traverses the complete terrain.

\subsubsection{Comparison Methods} We compared our ViSiR-Parkour with the following SOTA methods:

\noindent \textbf{Extreme Parkour (EP)} \cite{extremeParkour}: EP trains a teacher policy with PPO using privileged terrain observations, then distills the policy into a visual student policy using DAgger. EP uses the game-inspired curriculum (GIC) \cite{learningToWalkInMinutes}, which adjusts each robot's difficulty level according to traversal progress while keeping its assigned terrain type fixed. EP was selected as a representative visual parkour method that uses DAgger to distill a single privileged teacher policy into a depth-based student policy.

\noindent \textbf{START} \cite{START}: START uses a reconstruction network to predict a local heightmap from depth image observations. A locomotion policy uses the reconstructed heightmap and proprioceptive observations to generate robot joint actions. START jointly trains the reconstruction network and the locomotion policy in a single stage using PPO. GIC adjusts the terrain difficulty \cite{learningToWalkInMinutes}. START was selected as it is a single-stage visual locomotion method that explicitly reconstructs terrain geometry for foot placement on challenging terrains.

\noindent \textbf{Parkour in the Wild (PitW)} \cite{parkourInTheWild}: PitW uses a three-stage training framework. First, nine terrain-specific expert policies are trained with PPO using privileged observations. Second, the expert policies are distilled with DAgger into a single visual student policy. Third, the distilled student policy is fine-tuned using PPO, with new terrain types added to the training set through DRL fine-tuning. PitW uses GIC \cite{learningToWalkInMinutes} to adjust terrain difficulty during training. PitW was selected as it trains a single general visual parkour policy using multi-expert distillation followed by RL fine-tuning.

We extended EP, START and PitW to the A2 quadruped and trained them on the same terrains as ViSiR-Parkour. For PitW, we followed its original three-stage procedure for incorporating additional terrains: we trained and distilled the nine terrain-specific experts in the first two stages, then fine-tuned the student policy on the complete ViSiR-Parkour training terrain set in stage three.
\subsubsection{Procedure} For each method, we conducted three trials on each fixed terrain instance, resulting in 11,400 trials per method. The velocity command was randomly sampled in each trial. A trial terminated when the robot either reached 8 m of forward progress, fell, deviated from the evaluation terrains, or exceeded the allocated 30 s time limit. The time limit was selected based on the terrain length and velocity range to provide sufficient time for the robot to traverse the complete terrain.
\begin{figure}[!t]
\centering
\includegraphics[width=\columnwidth]{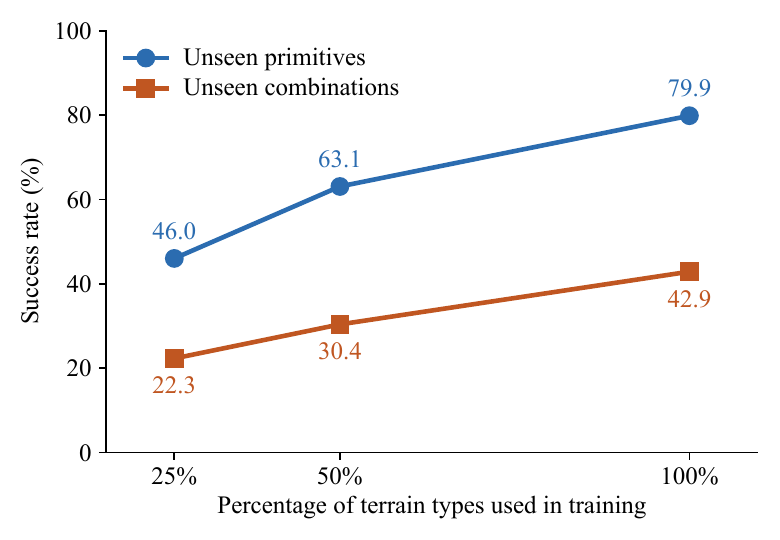}%
\caption{SR on unseen primitive and combination terrains for ViSiR-Parkour trained with 25\%, 50\%, and 100\% of the training terrain types.}
\label{fig:scaling_primitive_count}
\end{figure}

\subsubsection{Results} The SR results of ViSiR-Parkour and the comparison methods are presented in Table \ref{tab:simulated_comparison}. Overall, ViSiR-Parkour achieved the highest SR across all test terrains. EP had the lowest SR for in-distribution test terrains and the second lowest SR for out-of-distribution test terrains. ViSiR-Parkour achieved higher SR than EP as ViSiR-Parkour uses DRL to optimize task rewards during student training in addition to DAgger imitation learning, whereas EP trains its visual student policy only through DAgger to reproduce the teacher's actions. The perception gap between the teacher's privileged observations and the student's visual observations can introduce imitation errors that reduce the student's traversal success. ViSiR-Parkour achieved higher SR than START on both in-distribution and out-of-distribution terrains. START minimizes heightmap reconstruction errors without explicitly prioritizing regions needed for foot placement. Errors within the foot placement regions can cause the robot to misstep and fail to traverse the terrain. ViSiR-Parkour instead generates joint actions directly from visual and proprioceptive observations, avoiding terrain perception errors introduced by intermediate heightmap reconstruction. ViSiR-Parkour achieved higher SR than PitW on both in-distribution and out-of-distribution terrains as its PATR curriculum reallocates robots toward terrain types with low traversal success. This gives the parkour policy more training exposure to learn the locomotion skills needed to traverse those terrain types. In contrast, PitW's fixed terrain assignments during training can result in terrain types with low success rates being undertrained, resulting in lower SR. ViSiR-Parkour's higher SR on unseen primitive and combination terrains further demonstrates its generalizability to unseen terrain types and cross-terrain transitions. A statistically significant difference in SR among the four methods was found using a Friedman test ($\chi^2(3)=89.73$, $p$$<$$0.0001$). Post-hoc Wilcoxon signed-rank tests with Bonferroni correction found higher SRs for ViSiR-Parkour compared to all three methods: EP ($W=0$, $Z=-6.15$, $p$$<$$0.0001$), START ($W=17$, $Z=-5.99$, $p$$<0.0001$), and PitW ($W=45$, $Z=-5.65$, $p$$<0.0001$).

\subsection{Scalability Study}
A scalability study was conducted to investigate whether ViSiR-Parkour's generalizability to out-of-distribution terrains improves with the increase in the number of training terrain types. Three ViSiR-Parkour teacher policies were trained on 9, 18, and 37 primitive terrain types, corresponding to approximately 25\%, 50\%, and 100\% of the primitive training set. All three policies were trained for 10,000 iterations using the same hyperparameters and training procedure described in Section V-A. We evaluated the three policies on the same out-of-distribution primitive and combination terrains used in Section VI.A.

\subsubsection{Results} Fig. \ref{fig:scaling_primitive_count} shows the SR of the three policies on unseen primitive and combination terrains. Overall, ViSiR-Parkour's performance on out-of-distribution terrains improves as the number of terrain types used for training increases. This is due to training with more terrain types exposes the parkour policy to a wider range of terrain geometries and configurations, requiring it to adapt foot placement and body motion across different terrain traversal scenarios. As a result, the parkour policy learns locomotion skills that are able to generalize to unseen terrain geometries and transitions between different terrain types.

\begin{figure*}[!t]
\centering
\includegraphics[width=\textwidth]{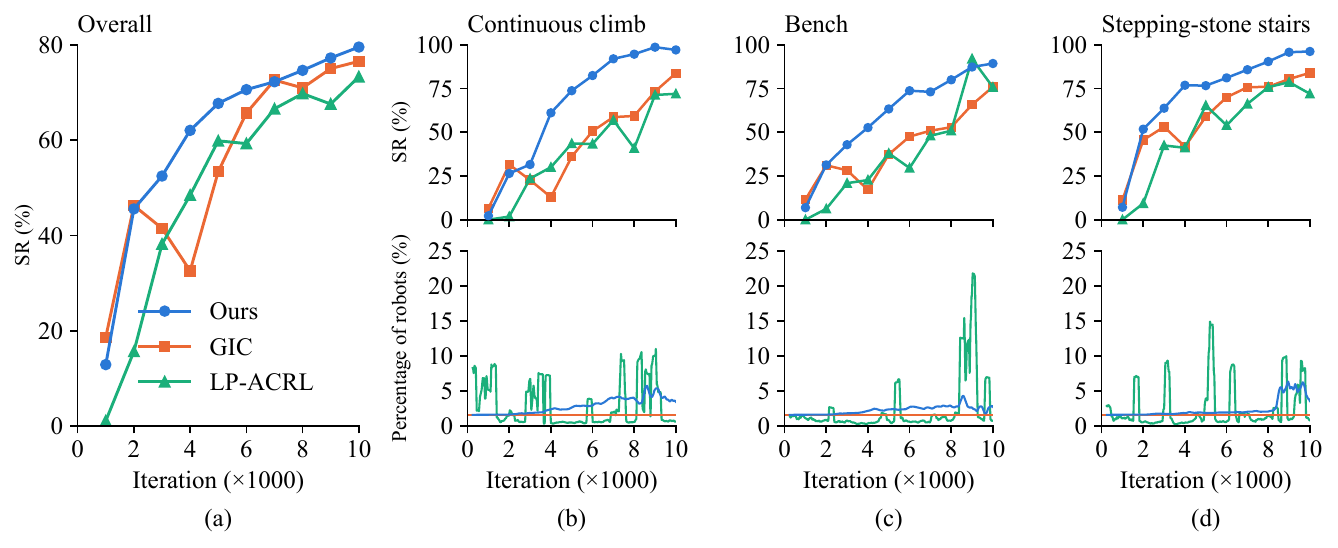}
\caption{Curriculum learning comparison between the PATR curriculum (Ours), the game-inspired curriculum (GIC), and LP-ACRL. (a): overall SR across all test terrains. (b)--(d) show SR (top) and the percentage of all training robots assigned to the corresponding terrain type during training (bottom) for three example terrain types: continuous climb, bench, and stepping-stone stairs, respectively.}
\label{fig:curriculum_ablation}
\end{figure*}

\subsection{Ablation Studies}
We conducted two sets of ablation studies to assess the impact of ViSiR-Parkour's main design choices in terms of the PATR curriculum learning method, and the effect of the hybrid policy learning method and the temporal feature aggregation in the student policy.

\subsubsection{Curriculum Learning}
We compared the proposed PATR curriculum learning method with two variants:

\noindent \textbf{ViSiR-Parkour w/ Game-Inspired Curriculum (GIC)}: This variant uses the game-inspired terrain curriculum in \cite{learningToWalkInMinutes}, in which each robot remains assigned to its initial terrain type and independently progresses through the terrain difficulty levels. GIC was selected as it is a standard and widely-used curriculum learning method for legged locomotion.

\noindent \textbf{ViSiR-Parkour w/ Learning Progress-based Automatic Curriculum Reinforcement Learning (LP-ACRL)}: This variant uses LP-ACRL \cite{LPACRL} as the curriculum learning method. It divides the locomotion task space into discrete tasks defined by terrain type, difficulty level, and velocity-command range. It estimates learning progress from changes in the mean episodic reward for each task and uses a softmax distribution to sample tasks with greater reward improvement more frequently. LP-ACRL was selected as it has demonstrated high sample efficiency for locomotion across diverse terrains.

\textit{Procedure:} We compared the ViSiR-Parkour with the two variants by evaluating their teacher policy checkpoints at intervals of 1,000 training iterations up to 10,000 iterations. Each checkpoint was evaluated on the same fixed terrain instances across all in-distribution and out-of-distribution terrains.

\textit{Results:} Fig. \ref{fig:curriculum_ablation} presents the SRs and robot allocations over training for the three methods. After 2,000 training iterations, ViSiR-Parkour generally achieved higher SR than both variants and reached comparable performance as the two variants in fewer training iterations, demonstrating greater sample efficiency. Fig. \ref{fig:curriculum_ablation}(b)--(d) show three challenging training terrain types: continuous climb, which consists of consecutive ascending and descending steps; bench, which requires climbing over seats and raised backrests; and stepping-stone stairs, which consist of individual stepping stones separated by gaps and arranged as ascending and descending steps.

On these terrains, GIC achieved lower SR than our PATR curriculum as GIC maintained a fixed number of robots on each terrain type, so challenging terrains receive no additional training even when their traversal success rates were low. LP-ACRL achieved lower SR than PATR. Its robot allocation exhibited sharp spikes followed by periods of low allocation. This is due to LP-ACRL allocating robots according to episodic reward improvement rather than traversal SR. When reward improvement rate on a specific terrain type becomes slower relative to other terrains, LP-ACRL can assign fewer robots to that terrain type even though its traversal SR remains low. This can also reduce the challenging terrain samples utilized during training before the policy has the opportunity to learn to reliably traverse them.

As the PATR curriculum learning method of ViSiR-Parkour reallocates robots according to terrain traversal success rates, it increases the number of robots on these challenging terrains during training. This provides more experience for the parkour policy to learn the necessary traversal skills, Fig. \ref{fig:curriculum_ablation}(b)-(d). As traversal performance improves, fewer additional robots are assigned to these terrain types, allowing training to focus on terrain types with lower success rates.

\begin{table}[!t]
\caption{Ablation Study of Hybrid Policy Learning and Temporal Feature Aggregation}
\label{tab:ablation_study}
\centering
\scriptsize
\setlength{\tabcolsep}{3pt}
\renewcommand{\arraystretch}{1.1}
\begin{tabular*}{\columnwidth}{@{\extracolsep{\fill}}lcc@{}}
\toprule
Variant & \shortstack{In-Distribution \\ SR (\%)} & \shortstack{Out-of-Distribution \\ SR (\%)} \\
\midrule
ViSiR-Parkour & 87.3 & 52.6 \\
ViSiR-Parkour w/o DRL & 84.0 & 42.8 \\
ViSiR-Parkour w/ DI & 73.1 & 40.7 \\
ViSiR-Parkour w/o GRU & 42.3 & 23.4 \\
\bottomrule
\end{tabular*}
\end{table}

\subsubsection{Hybrid Policy Learning and Temporal Feature Aggregation}
The following variants were evaluated against ViSiR-Parkour:

\noindent \textbf{ViSiR-Parkour w/o DRL}: The student policy was trained using only DAgger, without the PPO loss. This variant assesses the impact of PPO-based DRL on student policy training.

\noindent \textbf{ViSiR-Parkour w/ Decaying Imitation (DI)}: This variant gradually reduces the imitation-loss coefficient $\beta$ to zero during hybrid policy learning, after which the student policy is optimized using PPO alone \cite{realWorldHumanoidRL}. DI evaluates the contribution of the teacher supervision in the hybrid policy learning method.

\noindent \textbf{ViSiR-Parkour w/o GRU}: The student policy was implemented without GRU: the fused visual and proprioceptive features were mapped directly to the policy MLP. This variant evaluates the impact of temporal feature aggregation in the student policy.

\textit{Procedure:} We evaluated each variant on the same test terrains using the SR metric and the experimental procedure described in Section VI-A.

\textit{Results}: The SR results of ViSiR-Parkour and its variants are presented in Table \ref{tab:ablation_study}. ViSiR-Parkour achieved higher SR than all the variants on both the in-distribution and out-of-distribution terrains. Overall, ViSiR-Parkour w/o GRU achieved the lowest SR as removing the GRU prevented the student from retaining previously observed terrain information. This information is needed for robot foot placement where the terrain is outside the cameras' field of view, particularly of its hind feet. Without previous terrain information, the robot may miss the supporting surfaces and not be able to complete the traversal. ViSiR-Parkour w/o DRL had the second highest SR. With only the DAgger imitation learning, the perception gap between the teacher policy's privileged observations and the student policy's visual observations can cause errors in predicting the teacher's actions on unfamiliar terrain geometries. In ViSiR-Parkour, by additionally optimizing task rewards, DRL allows the student parkour policy to learn successful actions even when they differ from the teacher's, improving traversal of unseen terrains. ViSiR-Parkour w/ DI had lower SR than ViSiR-Parkour, as its performance declined during training due to the imitation weight approaching zero. Subsequent PPO training improved performance, however, did not surpass the level achieved before this decline, indicating the benefit of continued teacher supervision for preserving learned traversal skills. We also observed that with this variant, the robot lost the teacher's symmetric gait and often turned left or right when commanded only to move forward. Without the DAgger loss, the student policy training could no longer correct such deviations.

\begin{figure}[!t]
\centering
\includegraphics[width=\columnwidth]{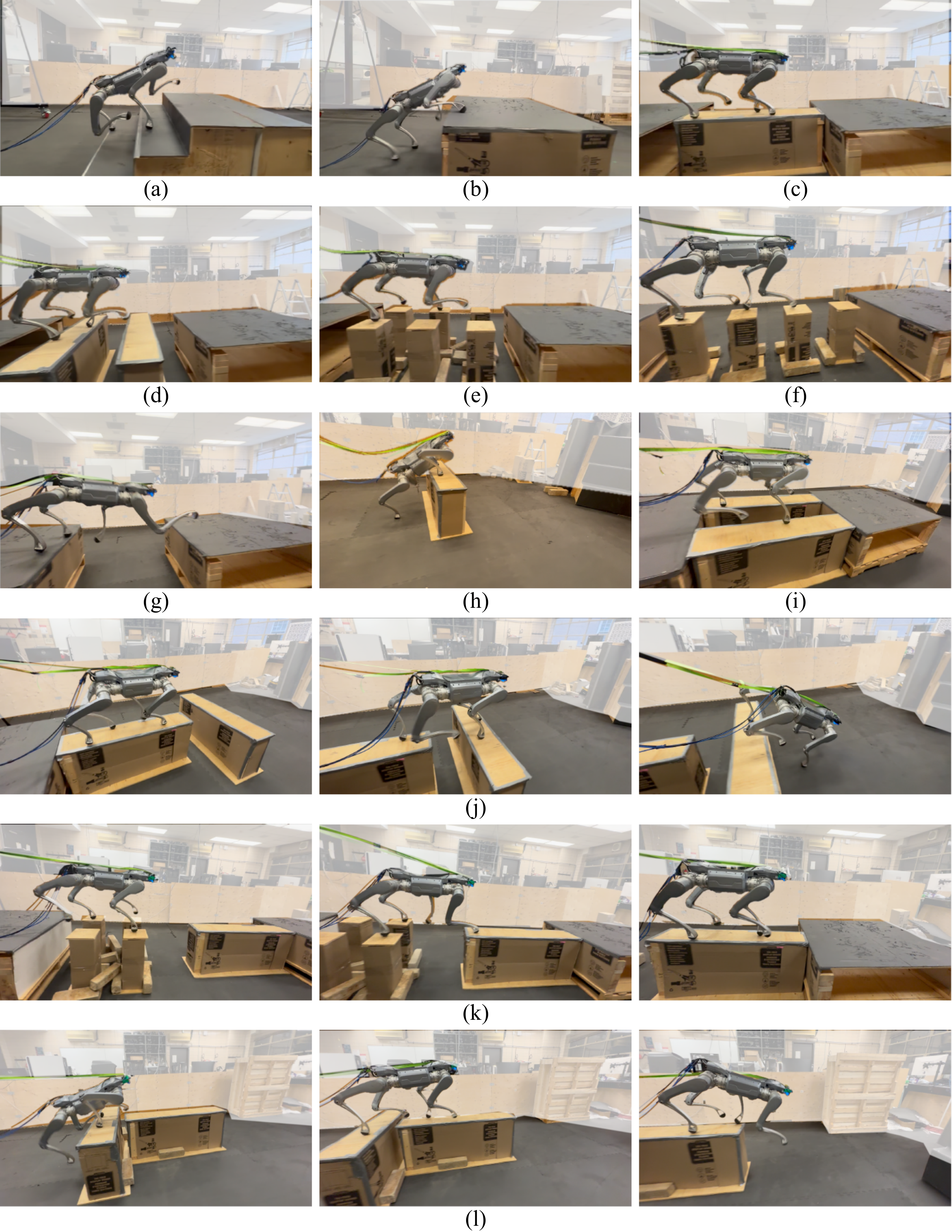}
\caption{Real-world terrains used in the quantitative comparison: (a) stairs, (b) platform climbing, (c) balance beams, (d) stepping beams, (e) stepping stones, (f) single-column stepping stones, (g) platform gaps, (h) hurdles, (i) parallel bars, and (j)--(l) three combination terrains.}
\label{fig:real_world_composed}
\end{figure}

\section{Real-World Experiments}
We evaluated ViSiR-Parkour in real-world experiments by conducting: 1) a quantitative comparison study of visual sim-to-real methods for legged robot parkour to assess real-world terrain traversal performance, and 2) numerous case studies to assess ViSiR-Parkour's generalizability across diverse indoor parkour courses and outdoor terrains in the wild, including unseen out-of-distribution terrain configurations.

\subsection{Comparison of Visual Sim-to-Real Methods}
We compared the performance of the proposed visual sim-to-real method of ViSiR-Parkour with a baseline method, denoted PitW*, that uses PitW's depth-noise corruption method for visual sim-to-real transfer \cite{parkourInTheWild}. We selected PitW's depth-noise augmentation approach for the baseline as it models artifacts in real-world stereo depth images during training to improve policy robustness to noisy depth observations. To ensure a fair comparison of visual sim-to-real transfer, PitW* uses the same teacher policy, student policy network, training terrains, hybrid policy learning procedure, and depth-observation delay model as ViSiR-Parkour. PitW* is trained with simulated depth images augmented with edge corruption, missing-depth patches, stereo blind spots, and Gaussian blur. Depth-observation delays are also randomized during training, following the procedure in \cite{parkourInTheWild}. During deployment, PitW* receives depth images from the onboard cameras and processes them using downsampling, depth clipping, and Gaussian blurring, while ViSiR-Parkour receives depth images estimated by LAS2.

\begin{table}[!t]
\caption{Real-World Comparison of Visual Sim-to-Real Methods}
\label{tab:real_world_sim2real_comparison}
\centering
\scriptsize
\setlength{\tabcolsep}{3pt}
\renewcommand{\arraystretch}{0.90}
\setlength{\aboverulesep}{0.5pt}
\setlength{\belowrulesep}{0.5pt}
\begin{tabular*}{\columnwidth}{@{\extracolsep{\fill}}llcc@{}}
\toprule
\multirow{2}{*}{Type} & \multirow{2}{*}{Terrain} & \multicolumn{2}{c}{SR (\%)} \\
\cmidrule(lr){3-4}
& & PitW* & ViSiR-Parkour \\
\midrule
\multirow{9}{*}{Primitive}
& Stairs & \textbf{100} & \textbf{100} \\
& Platform climbing & 60 & \textbf{80} \\
& Balance beams & \textbf{100} & \textbf{100} \\
& Stepping beams & 80 & \textbf{100} \\
& Stepping stones & 60 & \textbf{100} \\
& Single-column stepping stones & 0 & \textbf{80} \\
& Terrain Gaps & 20 & \textbf{60} \\
& Hurdles & 60 & \textbf{100} \\
& Parallel bars & \textbf{100} & \textbf{100} \\
\midrule
\multirow{3}{*}{Combination}
& Combination 1 & 40 & \textbf{80} \\
& Combination 2 & 20 & \textbf{80} \\
& Combination 3 & 20 & \textbf{80} \\
\midrule
\multicolumn{2}{l}{Overall} & 55.0 & \textbf{88.3} \\
\bottomrule
\end{tabular*}
\end{table}

\subsubsection{Procedure}
We deployed ViSiR-Parkour and PitW* on the A2 quadruped using the hardware set-up described in Section V-B. In each trial, PitW* received the raw depth images from the robot onboard stereo cameras, while ViSiR-Parkour received the depth images estimated by LAS2. The evaluation included nine primitive terrains and three combination terrains. The primitive terrains, Fig. \ref{fig:real_world_composed}(a)-(i), included stairs, platform climbing, balance beams, stepping beams, stepping stones, single-column stepping stones, terrain gaps, hurdles, and parallel bars. The combination terrains, Fig. \ref{fig:real_world_composed}(j)-(l), consisted of the following terrain types: 1) balance beam $\rightarrow$ stepping beams $\rightarrow$ climb down; 2) stepping stones $\rightarrow$ terrain gap $\rightarrow$ balance beam; and 3) hurdle $\rightarrow$ balance beam $\rightarrow$ climb down. For each method, we conducted five trials on each terrain.

\subsubsection{Results}
The SR results of ViSiR-Parkour and PitW* are presented in Table \ref{tab:real_world_sim2real_comparison}. Overall, ViSiR-Parkour achieved a higher SR than PitW*. Both methods achieved 100\% SR on stairs, balance beams, and parallel bars. Stair treads allow variation in lateral foot placement, while balance beams and parallel bars allow variation along their length. With PitW*, the robot's feet can still land on these supporting surfaces when noisy depth observations cause placement errors in these directions. ViSiR-Parkour achieved higher SR than PitW* on the remaining primitive terrains and all combination terrains. This is due to noise and missing depth values in PitW*'s depth observations which can lead the robot to misjudge terrain edges and obstacle heights, resulting in missed footholds and collisions. In contrast, LAS2 provided ViSiR-Parkour with clean depth estimates of the terrain geometry, enabling accurate foot placement over these challenging terrains. A Wilcoxon signed-rank test found that ViSiR-Parkour had a statistically significant difference in SR between ViSiR-Parkour and PitW* ($W=0$, $p=0.0039$).

\begin{figure*}[!t]
\centering
\includegraphics[width=0.96\textwidth]{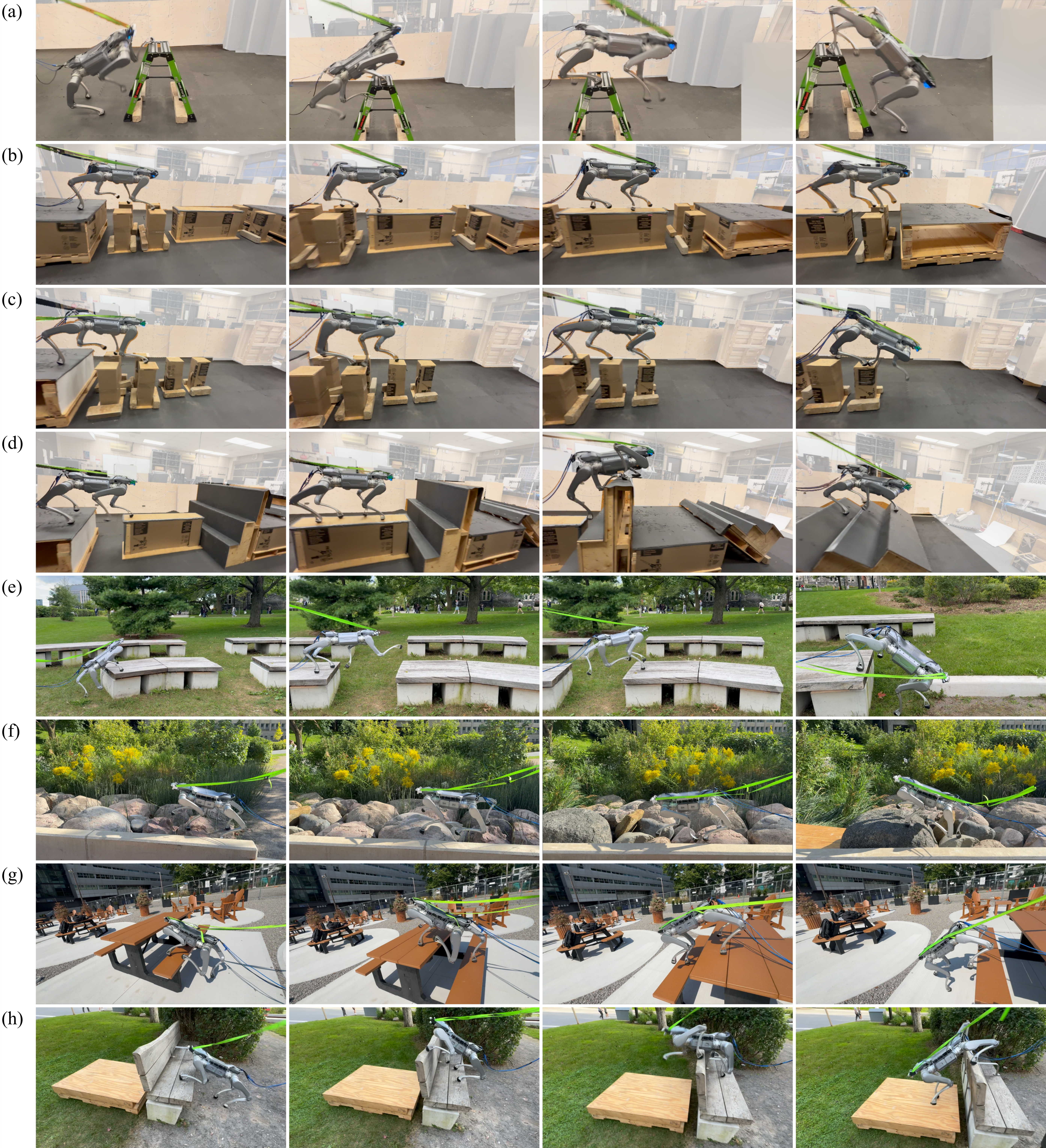}
\caption{Time-ordered frames from eight real-world traversals ordered chronologically from left to right. (a) traversal over an indoor step ladder. (b)--(d) three indoor combination terrains. (e)--(h) a gap between bench segments, a rock pile, a picnic table, and a bench with a backrest, respectively.}
\label{fig:real_world_motion}
\end{figure*}

\subsection{Case Studies on Diverse Real-World Terrains}

We conducted case studies to investigate the generalizability of ViSiR-Parkour across diverse challenging terrains in indoor parkour courses and outdoor in-the-wild environments. A video of the real-world case studies is provided at our project webpage: \url{https://visir-parkour.github.io}.

\subsubsection{Case Study 1 Indoor Terrains}
\noindent The ability of the A2-embedded with ViSiR-Parkour to traverse four challenging indoor terrains was investigated. In Fig. \ref{fig:real_world_motion}(a), the robot successfully climbed over a stepladder with steps that are narrower than the A2's overall standing width, restricting lateral foot placement. The robot then moved its body over the top and descended on the opposite side. In Fig. \ref{fig:real_world_motion}(b), the robot successfully traversed a combination terrain of stepping stones, a balance beam, and another set of stepping stones. The stepping stones constrained foot placement in both the lateral and forward directions, while the narrow balance beam constrained lateral foot placement. During each transition, the robot's front and hind feet contacted different terrain types, requiring coordinated foot placement and body motion to maintain balance. In Fig. \ref{fig:real_world_motion}(c), the robot successfully traversed two rows of two-column stepping stones followed by single-column stepping stones and descended to the floor. This transition required both the left and right feet to land within the same narrow column of stepping stones, imposing tighter lateral foot-placement constraints than the preceding two-column arrangement. The final descent required the robot to control its body pitch and maintain balance as its front feet reached the floor while its hind feet remained on the elevated stepping stones. In Fig. \ref{fig:real_world_motion}(d), the robot successfully crossed a 0.6 m gap from a platform onto a balance beam. This required sufficient forward displacement to clear the gap and accurate lateral foot placement to land on the narrow beam. The robot then traversed the balance beam, climbed over a vertically positioned staircase onto another platform, and descended a sloped staircase to the floor. Both staircase configurations were unseen during training, requiring accurate foot placement on the narrow upper surface of the vertically positioned staircase and control of body pitch to avoid falling forward when descending the sloped staircase. These successful traversals demonstrate ViSiR-Parkour's generalizability from simulated training terrains to real-world indoor terrains, including unseen terrains and combination terrains requiring transitions between terrain types.

\subsubsection{Case Study 2 Outdoor Terrains}
\noindent The generalizability of the A2-embedded with ViSiR-Parkour to traverse four challenging outdoor terrains was investigated. Figure \ref{fig:real_world_motion}(e) shows the robot successfully climbing onto the first bench segment, jumping across a 1 m gap to the second segment, and descending to the ground. Crossing the gap required accurate foot placement near the edge of the first bench segment before takeoff, sufficient forward displacement to clear the gap, and accurate foot placement on the second bench segment upon landing. The rock pile in Fig. \ref{fig:real_world_motion}(f) contained rocks with irregular surfaces at varying heights and inclinations. Gaps between the rocks restricted the available footholds, while the uneven surfaces required adjustments in foot placement and body orientation to prevent missed footholds and loss of balance. The robot successfully traversed the pile by stepping on individual rock surfaces. In Fig. \ref{fig:real_world_motion}(g), the robot successfully climbed onto the picnic-table seat to step up to the picnic table, traversed the table, and descended on the opposite side. The narrower seat (than table) constrained hind-foot placement as the robot raised its front feet onto the table. Descent required control of body pitch to avoid falling forward as the front feet reached the ground while the hind feet remained on the seat. In Fig. \ref{fig:real_world_motion}(h), the robot first climbed onto a bench seat. It then used the upper edge of the backrest as a foothold to raise its body, while its hind feet remained supported on the seat. The robot subsequently climbed over the backrest and descended on the opposite side. The narrow upper edge of the backrest limited the area available for foot placement, while climbing over the backrest required sufficient body and hind-leg clearance to avoid collisions. These successful traversals demonstrate ViSiR-Parkour's generalizability to outdoor terrains in the wild through coordinated foot placement and body motion across different terrains and terrain transitions.

The visual sim-to-real performance of ViSiR-Parkour is dependent on the accuracy of stereo depth estimation. Significant changes in illumination can produce depth-estimation errors in LAS2 \cite{LAS2}, potentially creating discrepancies from the clean simulated depth observations used to train the student policy. This could be addressed by incorporating active illumination on the robot to improve depth estimation under low-lighting conditions \cite{realsenseIllumination}.

\section{Conclusion}
In this paper, we presented Visual Sim-to-Real Parkour (ViSiR-Parkour), a novel locomotion architecture that addresses the visual sim-to-real and terrain generalization challenges of legged robot parkour in the wild. Our architecture trains the teacher and student policies in two stages. In the first stage, a teacher policy is trained using PPO with privileged observations. We scale terrain diversity and use the progress-aware terrain reallocation curriculum to train the teacher policy across varied primitive and combination terrains. In the second stage, the visual student policy is trained through hybrid policy learning using DAgger and PPO. During real-world deployment, a learning-based stereo matching model generates clean depth observations to reduce the visual sim-to-real gap. Extensive simulated experiments showed that ViSiR-Parkour traverses diverse challenging courses and generalizes to terrain geometries and layouts excluded from training. The ablation study further verified the contribution of the main architecture components. Real-world experiments verified the proposed visual sim-to-real method's ability to enable a quadruped to successfully traverse indoor parkour courses and challenging outdoor terrains. Future work will investigate the application and performance of ViSiR-Parkour on other legged robot embodiments, such as humanoid robots.

\section*{Acknowledgment}
The authors would like to thank the following members of the ASBLab at UofT for their assistance in conducting the experiments: Glenn Takashi Shimoda, Georgia Jovanovic, and Zidong Han.

\renewcommand{\refname}{References}
\bibliographystyle{IEEEtran}
\bibliography{references}

@IEEEtranBSTCTL{IEEEReferenceAuthorControl,
  CTLuse_forced_etal = {yes},
  CTLmax_names_forced_etal = {6},
  CTLnames_show_etal = {1},
}

@article{learningAgileDynamicMotorSkills,
  author = {Hwangbo, Jemin and Lee, Joonho and Dosovitskiy, Alexey and Bellicoso, Dario and Tsounis, Vassilios and Koltun, Vladlen and Hutter, Marco},
  title = {{Learning agile and dynamic motor skills for legged robots}},
  journal = {Sci. Robot.},
  year = {2019},
  volume = {4},
  number = {26},
  pages = {eaau5872},
}

@inproceedings{dynamicsRandomization,
  author = {Peng, Xue Bin and Andrychowicz, Marcin and Zaremba, Wojciech and Abbeel, Pieter},
  title = {{Sim-to-Real Transfer of Robotic Control with Dynamics Randomization}},
  booktitle = {Proc. IEEE Int. Conf. Robot. Automat.},
  year = {2018},
  pages = {3803--3810},
  doi = {10.1109/ICRA.2018.8460528},
}

@inproceedings{domainRandomization,
  author = {Tobin, Josh and Fong, Rachel and Ray, Alex and Schneider, Jonas and Zaremba, Wojciech and Abbeel, Pieter},
  title = {{Domain Randomization for Transferring Deep Neural Networks from Simulation to the Real World}},
  booktitle = {Proc. IEEE/RSJ Int. Conf. Intell. Robots Syst.},
  year = {2017},
  pages = {23--30},
  doi = {10.1109/IROS.2017.8202133},
}

@article{handDrawnMapNavigation,
  author = {Tan, Aaron Hao and Fung, Angus and Wang, Haitong and Nejat, Goldie},
  title = {{Mobile Robot Navigation Using Hand-Drawn Maps: A Vision Language Model Approach}},
  journal = {IEEE Robot. Autom. Lett.},
  year = {2025},
  volume = {10},
  number = {7},
  pages = {7667--7674},
}

@article{xNav,
  author = {Wang, Haitong and Tan, Aaron Hao and Fung, Angus and Nejat, Goldie},
  title = {{X-Nav: Learning End-to-End Cross-Embodiment Navigation for Mobile Robots}},
  journal = {IEEE Robot. Autom. Lett.},
  year = {2026},
  volume = {11},
  number = {1},
  pages = {698--705},
}

@article{egoNav,
  author = {Wang, Weizhuo and Ze, Yanjie and Liu, C. Karen and Kennedy, III, Monroe},
  title = {{Learning Humanoid Navigation From Human Data}},
  journal = {IEEE Robot. Autom. Lett.},
  volume = {11},
  number = {9},
  pages = {10449--10456},
  month = sep,
  year = {2026},
  doi = {10.1109/LRA.2026.3710346},
}

@article{mineTunnelExploration,
  author = {Miller, Ian D. and Cladera, Fernando and Cowley, Anthony and Shivakumar, Shreyas S. and Lee, Elijah S. and Jarin-Lipschitz, Laura and Bhat, Akhilesh and Rodrigues, Neil and Zhou, Alex and Cohen, Avraham and Kulkarni, Adarsh and Laney, James and Taylor, Camillo Jose and Kumar, Vijay},
  title = {{Mine Tunnel Exploration Using Multiple Quadrupedal Robots}},
  journal = {IEEE Robot. Autom. Lett.},
  year = {2020},
  volume = {5},
  number = {2},
  pages = {2840--2847},
}

@article{traversabilityAwareExploration,
  author = {Prágr, Miloš and Bayer, Jan and Faigl, Jan},
  title = {{Autonomous Robotic Exploration with Simultaneous Environment and Traversability Models Learning}},
  journal = {Front. Robot. AI},
  year = {2022},
  volume = {9},
  pages = {910113},
}

@inproceedings{umiOnLegs,
  author = {Ha, Huy and Gao, Yihuai and Fu, Zipeng and Tan, Jie and Song, Shuran},
  title = {{UMI-on-Legs: Making Manipulation Policies Mobile with Manipulation-Centric Whole-Body Controllers}},
  booktitle = {Proc. Conf. Robot Learn.},
  year = {2025},
  volume = {270},
  pages = {5254--5270},
}

@inproceedings{humanPlus,
  author = {Fu, Zipeng and Zhao, Qingqing and Wu, Qi and Wetzstein, Gordon and Finn, Chelsea},
  title = {{HumanPlus: Humanoid Shadowing and Imitation from Humans}},
  booktitle = {Proc. Conf. Robot Learn.},
  year = {2025},
  volume = {270},
  pages = {2828--2844},
}

@article{embodiedAISurvey,
  author = {Lisondra, Matthew and Benhabib, Beno and Nejat, Goldie},
  title = {{Embodied AI with Foundation Models for Mobile Service Robots: A Systematic Review}},
  journal = {Robotics},
  year = {2026},
  volume = {15},
  number = {3},
  pages = {55},
  doi = {10.3390/robotics15030055},
}

@article{anymalParkour,
  author = {Hoeller, David and Rudin, Nikita and Sako, Dhionis and Hutter, Marco},
  title = {{ANYmal parkour: Learning agile navigation for quadrupedal robots}},
  journal = {Sci. Robot.},
  year = {2024},
  volume = {9},
  number = {88},
  pages = {eadi7566},
}

@inproceedings{robotParkourLearning,
  author = {Zhuang, Ziwen and Fu, Zipeng and Wang, Jianren and Atkeson, Christopher and Schwertfeger, Soeren and Finn, Chelsea and Zhao, Hang},
  title = {{Robot Parkour Learning}},
  booktitle = {Proc. Conf. Robot Learn.},
  year = {2023},
  volume = {229},
  pages = {73--92},
}

@article{parkourInTheWild,
  author = {Rudin, Nikita and He, Junzhe and Aurand, Joshua and Hutter, Marco},
  title = {{Parkour in the Wild: Learning a General and Extensible Agile Locomotion Policy Using Multi-expert Distillation and RL Fine-tuning}},
  journal = {Int. J. Robot. Res.},
  year = {2026},
}

@misc{PPO,
  author = {Schulman, John and Wolski, Filip and Dhariwal, Prafulla and Radford, Alec and Klimov, Oleg},
  title = {{Proximal Policy Optimization Algorithms}},
  year = {2017},
  note = {{\emph{arXiv}}:1707.06347},
}

@inproceedings{extremeParkour,
  author = {Cheng, Xuxin and Shi, Kexin and Agarwal, Ananye and Pathak, Deepak},
  title = {{Extreme Parkour with Legged Robots}},
  booktitle = {Proc. IEEE Int. Conf. Robot. Automat.},
  year = {2024},
  pages = {11443--11450},
}

@article{challengingTerrainLocomotion,
  author = {Lee, Joonho and Hwangbo, Jemin and Wellhausen, Lorenz and Koltun, Vladlen and Hutter, Marco},
  title = {{Learning quadrupedal locomotion over challenging terrain}},
  journal = {Sci. Robot.},
  year = {2020},
  volume = {5},
  number = {47},
  pages = {eabc5986},
}

@inproceedings{dynamicsRandomizationRevisited,
  author = {Xie, Zhaoming and Da, Xingye and van de Panne, Michiel and Babich, Buck and Garg, Animesh},
  title = {{Dynamics Randomization Revisited: A Case Study for Quadrupedal Locomotion}},
  booktitle = {Proc. IEEE Int. Conf. Robot. Automat.},
  year = {2021},
  pages = {4955-4961},
}

@inproceedings{locoFormer,
  author = {Liu, Min and Pathak, Deepak and Agarwal, Ananye},
  title = {{LocoFormer: Generalist Locomotion via Long-context Adaptation}},
  booktitle = {Proc. Conf. Robot Learn.},
  year = {2025},
  volume = {305},
  pages = {532--546},
}

@inproceedings{simToRealQuadrupedLocomotion,
  author = {Tan, Jie and Zhang, Tingnan and Coumans, Erwin and Iscen, Atil and Bai, Yunfei and Hafner, Danijar and Bohez, Steven and Vanhoucke, Vincent},
  title = {{Sim-to-Real: Learning Agile Locomotion For Quadruped Robots}},
  booktitle = {Proc. Robot. Sci. Syst.},
  year = {2018},
}

@article{beyondMimic,
  author = {Liao, Qiayuan and Truong, Takara E. and Huang, Xiaoyu and Gao, Yuman and Tevet, Guy and Sreenath, Koushil and Liu, C. Karen},
  title = {{BeyondMimic: From Motion Tracking to Versatile Humanoid Control via Guided Diffusion}},
  journal = {Sci. Robot.},
  volume = {11},
  number = {117},
  pages = {eadx8924},
  year = {2026},
}

@article{PACE,
  author = {Bjelonic, Filip and Tischhauser, Fabian and Hutter, Marco},
  title = {{Towards bridging the gap: Systematic sim-to-real transfer for diverse legged robots}},
  journal = {Int. J. Robot. Res.},
  year = {2026},
}

@inproceedings{ASAP,
  author = {He, Tairan and Gao, Jiawei and Xiao, Wenli and Zhang, Yuanhang and Wang, Zi and Wang, Jiashun and Luo, Zhengyi and He, Guanqi and Sobanbabu, Nikhil and Pan, Chaoyi and Yi, Zeji and Qu, Guannan and Kitani, Kris and Hodgins, Jessica K. and Fan, Linxi and Zhu, Yuke and Liu, Changliu and Shi, Guanya},
  title = {{ASAP: Aligning Simulation and Real-World Physics for Learning Agile Humanoid Whole-Body Skills}},
  booktitle = {Proc. Robot. Sci. Syst.},
  year = {2025},
}

@inproceedings{athleticLocoManipulationSimToReal,
  author = {Fey, Nolan and Margolis, Gabriel B. and Peticco, Martin and Agrawal, Pulkit},
  title = {{Bridging the Sim-to-Real Gap for Athletic Loco-Manipulation}},
  booktitle = {Proc. Robot. Sci. Syst.},
  year = {2025},
}

@inproceedings{RMA,
  author = {Kumar, Ashish and Fu, Zipeng and Pathak, Deepak and Malik, Jitendra},
  title = {{RMA: Rapid Motor Adaptation for Legged Robots}},
  booktitle = {Proc. Robot. Sci. Syst.},
  year = {2021},
}

@inproceedings{RL2AC,
  author = {Lyu, Shangke and Lang, Xin and Zhao, Han and Zhang, Hongyin and Ding, Pengxiang and Wang, Donglin},
  title = {{RL2AC: Reinforcement Learning-based Rapid Online Adaptive Control for Legged Robot Robust Locomotion}},
  booktitle = {Proc. Robot. Sci. Syst.},
  year = {2024},
}

@inproceedings{visualLocomotion,
  author = {Yu, Wenhao and Jain, Deepali and Escontrela, Alejandro and Iscen, Atil and Xu, Peng and Coumans, Erwin and Ha, Sehoon and Tan, Jie and Zhang, Tingnan},
  title = {{Visual-Locomotion: Learning to Walk on Complex Terrains with Vision}},
  booktitle = {Proc. Conf. Robot Learn.},
  year = {2022},
  volume = {164},
  pages = {1291--1302},
}

@inproceedings{nowYouSeeThat,
  author = {Sun, Wandong and Su, Yongbo and Huang, Leoric and Zhang, Alex and Wei, Dwyane and San, Mu and Tian, Daniel and Cao, Ellie and Yan, Finn and Xie, Ethan and Xie, Zongwu},
  title = {{Now You See That: Learning End-to-End Humanoid Locomotion from Raw Pixels}},
  booktitle = {Proc. Robot. Sci. Syst.},
  year = {2026},
}

@misc{hikingInTheWild,
  author = {Zhu, Shaoting and Zhuang, Ziwen and Zhao, Mengjie and Lee, Kun-Ying and Zhao, Hang},
  title = {{Hiking in the Wild: A Scalable Perceptive Parkour Framework for Humanoids}},
  year = {2026},
  note = {{\emph{arXiv}}:2601.07718},
}

@inproceedings{visualParkourGeneratedImages,
  author = {Yu, Alan and Yang, Ge and Choi, Ran and Ravan, Yajvan and Leonard, John and Isola, Phillip},
  title = {{Learning Visual Parkour from Generated Images}},
  booktitle = {Proc. Conf. Robot Learn.},
  year = {2025},
  volume = {270},
  pages = {2500--2516},
}

@inproceedings{VIRAL,
  author = {He, Tairan and Wang, Zi and Xue, Haoru and Ben, Qingwei and Luo, Zhengyi and Xiao, Wenli and Yuan, Ye and Da, Xingye and Castañeda, Fernando and Sastry, Shankar and Liu, Changliu and Shi, Guanya and Fan, Linxi and Zhu, Yuke},
  title = {{VIRAL: Visual Sim-to-Real at Scale for Humanoid Loco-Manipulation}},
  booktitle = {Proc. IEEE/CVF Conf. Comput. Vis. Pattern Recognit.},
  year = {2026},
  pages = {13430--13441},
}

@inproceedings{doorMan,
  author = {Xue, Haoru and He, Tairan and Wang, Zi and Ben, Qingwei and Xiao, Wenli and Luo, Zhengyi and Da, Xingye and Castañeda, Fernando and Shi, Guanya and Sastry, Shankar and Fan, Linxi "Jim" and Zhu, Yuke},
  title = {{Opening the Sim-to-Real Door for Humanoid Pixel-to-Action Policy Transfer}},
  booktitle = {Proc. IEEE/CVF Conf. Comput. Vis. Pattern Recognit.},
  year = {2026},
  pages = {6642--6652},
}

@article{PIE,
  author = {Luo, Shixin and Li, Songbo and Yu, Ruiqi and Wang, Zhicheng and Wu, Jun and Zhu, Qiuguo},
  title = {{PIE: Parkour with Implicit-Explicit Learning Framework for Legged Robots}},
  journal = {IEEE Robot. Autom. Lett.},
  year = {2024},
  volume = {9},
  number = {11},
  pages = {9986--9993},
}

@article{START,
  author = {Yu, Ruiqi and Wang, Qianshi and Li, Hongyi and Jun, Zheng and Wang, Zhicheng and Wu, Jun and Zhu, Qiuguo},
  title = {{START: Traversing Sparse Footholds with Terrain Reconstruction}},
  journal = {IEEE Robot. Autom. Lett.},
  year = {2026},
  volume = {11},
  number = {2},
  pages = {2194--2201},
  doi = {10.1109/LRA.2025.3645649},
}

@inproceedings{robustLadderClimbing,
  author = {Vogel, Dylan and Baines, Robert and Church, Joseph and Lotzer, Julian and Werner, Karl and Hutter, Marco},
  title = {{Robust Ladder Climbing with a Quadrupedal Robot}},
  booktitle = {Proc. IEEE/RSJ Int. Conf. Intell. Robots Syst.},
  year = {2025},
  pages = {7239--7244},
}

@misc{stairMaster,
  author = {Tang, Xincheng and Xie, Youhan and Shu, Zhengjie and Li, Wanyu and Jiang, Lai and Hu, Wenkang and Li, Yitong and Yang, Ruigang},
  title = {{StairMaster: Learning to Conquer Risky Hollow Stairs for Agile Quadrupedal Robots}},
  year = {2026},
  note = {{\emph{arXiv}}:2606.25765},
}

@misc{walkThePLANC,
  author = {Dai, Min and Compton, William D. and Li, Junheng and Yang, Lizhi and Ames, Aaron D.},
  title = {{Walk the PLANC: Physics-Guided RL for Agile Humanoid Locomotion on Constrained Footholds}},
  year = {2026},
  note = {{\emph{arXiv}}:2601.06286},
}

@article{AME,
  author = {He, Junzhe and Zhang, Chong and Jenelten, Fabian and Grandia, Ruben and Bächer, Moritz and Hutter, Marco},
  title = {{Attention-based map encoding for learning generalized legged locomotion}},
  journal = {Sci. Robot.},
  year = {2025},
  volume = {10},
  number = {105},
  pages = {eadv3604},
}

@article{AME2,
  author = {Zhang, Chong and Klemm, Victor and Yang, Fan and Hutter, Marco},
  title = {{Agile and Generalized Legged Locomotion via Attention-Based Neural Map Encoding}},
  journal = {IEEE Trans. Robot.},
  volume = {42},
  pages = {4079--4098},
  year = {2026},
}

@inproceedings{hamlet,
  author = {Wang, Haochen and Shi, Zhiwei and Zhu, Chengxi and Qiao, Yafei and Zhang, Cheng and Yang, Fan and Ren, Pengjie and Lu, Lan and Xuan, Dong},
  title = {{Integrating Learning-Based Manipulation and Physics-Based Locomotion for Whole-Body Badminton Robot Control}},
  booktitle = {Proc. IEEE Int. Conf. Robot. Automat.},
  year = {2025},
  pages = {15114--15120},
}

@misc{TTTParkour,
  author = {Zhu, Shaoting and Ye, Baijun and Wang, Jiaxuan and Chen, Jiakang and Zhuang, Ziwen and Mou, Linzhan and Huang, Runhan and Zhao, Hang},
  title = {{TTT-Parkour: Rapid Test-Time Training for Perceptive Robot Parkour}},
  year = {2026},
  note = {{\emph{arXiv}}:2602.02331},
}

@inproceedings{terrainRobustnessBenchmark,
  author = {Zhang, Chong and Yang, Lizhi},
  title = {{Generating a Terrain-Robustness Benchmark for Legged Locomotion: A Prototype via Terrain Authoring and Active Learning}},
  booktitle = {Proc. IEEE Int. Conf. Robot. Automat.},
  year = {2023},
  pages = {9190--9196},
}

@inproceedings{diverseChallengingTerrains,
  author = {Esquerre-Pourtère, Arthur and Kim, Minsoo and Park, Jaeheung},
  title = {{Generating Diverse Challenging Terrains for Legged Robots Using Quality-Diversity Algorithm}},
  booktitle = {Proc. IEEE Int. Conf. Robot. Automat.},
  year = {2025},
  pages = {7654-7660},
}

@article{agilePerceptiveMultiskillLocomotion,
  author = {Kang, Jun-Gill and Park, Jaehyun and Song, Tae-Gyu and Kim, Joon-Ha and Hong, Seungwoo and Park, Hae-Won},
  title = {{Agile Perceptive Multiskill Locomotion for Quadrupedal Robots in the Wild}},
  journal = {Sci. Robot.},
  year = {2026},
  volume = {11},
  number = {116},
  pages = {eadz7397},
  doi = {10.1126/scirobotics.adz7397},
}

@misc{LAS2,
  author = {Jing, Junpeng and Zuo, Ronglai and Shen, Zhelun and Zhou, Shangchen and Potamias, Rolandos Alexandros and Zafeiriou, Stefanos and Mikolajczyk, Krystian and Deng, Jiankang},
  title = {{Lite Any Stereo V2: Faster and Stronger Efficient Zero-Shot Stereo Matching}},
  year = {2026},
  note = {{\emph{arXiv}}:2606.24457},
}

@misc{ONNX,
  author = {Bai, Junjie and Lu, Fang and Zhang, Ke and others},
  title = {{ONNX: Open Neural Network Exchange}},
  year = {2019},
  howpublished = {\url{https://onnx.ai/}},
}

@manual{TensorRT,
  author = {{NVIDIA Corporation}},
  title = {{NVIDIA TensorRT Developer Guide}},
  year = {2024},
  organization = {NVIDIA Corporation},
  url = {https://docs.nvidia.com/deeplearning/tensorrt/},
}

@misc{ladderMan,
  author = {Zhao, Siheng and Zhang, Yuanhang and Lu, Ziqi and Abbeel, Pieter and Duan, Rocky and Sreenath, Koushil and Wang, Yue and Liu, C. Karen and Shi, Guanya},
  title = {{LadderMan: Learning Humanoid Perceptive Ladder Climbing}},
  year = {2026},
  note = {{\emph{arXiv}}:2606.05873},
}

@article{deepTrackingControl,
  author = {Jenelten, Fabian and He, Junzhe and Farshidian, Farbod and Hutter, Marco},
  title = {{DTC: Deep Tracking Control}},
  journal = {Sci. Robot.},
  year = {2024},
  volume = {9},
  number = {86},
  pages = {eadh5401},
}

@article{highSpeedControlNavigation,
  author = {Kim, Hyeongjun and Oh, Hyunsik and Park, Jeongsoo and Kim, Yunho and Youm, Donghoon and Jung, Moonkyu and Lee, Minho and Hwangbo, Jemin},
  title = {{High-speed control and navigation for quadrupedal robots on complex and discrete terrain}},
  journal = {Sci. Robot.},
  year = {2025},
  volume = {10},
  number = {102},
  pages = {eads6192},
}

@inproceedings{learningToWalkInMinutes,
  author = {Rudin, Nikita and Hoeller, David and Reist, Philipp and Hutter, Marco},
  title = {{Learning to Walk in Minutes Using Massively Parallel Deep Reinforcement Learning}},
  booktitle = {Proc. Conf. Robot Learn.},
  year = {2022},
  volume = {164},
  pages = {91--100},
}

@inproceedings{asymmetricActorCritic,
  author = {Pinto, Lerrel and Andrychowicz, Marcin and Welinder, Peter and Zaremba, Wojciech and Abbeel, Pieter},
  title = {{Asymmetric Actor Critic for Image-Based Robot Learning}},
  booktitle = {Proc. Robot. Sci. Syst.},
  year = {2018},
}

@inproceedings{symmetryRobotPolicies,
  author = {Mittal, Mayank and Rudin, Nikita and Klemm, Victor and Allshire, Arthur and Hutter, Marco},
  title = {{Symmetry Considerations for Learning Task Symmetric Robot Policies}},
  booktitle = {Proc. IEEE Int. Conf. Robot. Automat.},
  year = {2024},
  pages = {7433--7439},
}

@inproceedings{gatedRecurrentUnit,
  author = {Cho, Kyunghyun and van Merri\"enboer, Bart and Gulcehre, Caglar and Bahdanau, Dzmitry and Bougares, Fethi and Schwenk, Holger and Bengio, Yoshua},
  title = {{Learning Phrase Representations Using RNN Encoder--Decoder for Statistical Machine Translation}},
  booktitle = {Proc. Conf. Empirical Methods Natural Lang. Process.},
  year = {2014},
  pages = {1724--1734},
}

@inproceedings{DAgger,
  author = {Ross, St{\'e}phane and Gordon, Geoffrey J. and Bagnell, J. Andrew},
  title = {{A Reduction of Imitation Learning and Structured Prediction to No-Regret Online Learning}},
  booktitle = {Proc. Int. Conf. Artif. Intell. Stat.},
  year = {2011},
  volume = {15},
  pages = {627--635},
}

@misc{isaacLab,
  author = {Mayank Mittal and others},
  title = {{Isaac Lab}: A {GPU}-Accelerated Simulation Framework for Multi-Modal Robot Learning},
  year = {2025},
  note = {{\emph{arXiv}}:2511.04831},
}

@inproceedings{adam,
  author = {Kingma, Diederik P. and Ba, Jimmy},
  title = {{Adam: A Method for Stochastic Optimization}},
  booktitle = {Proc. Int. Conf. Learn. Represent.},
  year = {2015},
}

@article{LPACRL,
  author = {Li, Ziming and Li, Chenhao and Hutter, Marco},
  title = {{Scaling Rough Terrain Locomotion With Automatic Curriculum Reinforcement Learning}},
  journal = {IEEE Robot. Autom. Lett.},
  year = {2026},
  volume = {11},
  number = {8},
  pages = {9295--9302},
}

@article{realWorldHumanoidRL,
  author = {Radosavovic, Ilija and Xiao, Tete and Zhang, Bike and Darrell, Trevor and Malik, Jitendra and Sreenath, Koushil},
  title = {{Real-World Humanoid Locomotion with Reinforcement Learning}},
  journal = {Sci. Robot.},
  year = {2024},
  volume = {9},
  number = {89},
  pages = {eadi9579},
}

@techreport{realsenseIllumination,
  author = {Grunnet-Jepsen, Anders and Sweetser, John N. and Winer, Paul and Takagi, Akihiro and Woodfill, John},
  title = {{Projectors for Intel RealSense Depth Cameras D4xx}},
  institution = {Intel Corp.},
  type = {White Paper},
  number = {rev. 1.0},
  year = {2018},
}

\end{document}